\documentclass[%
 reprint,
 aps,
 pre,
 showkeys,
 floatfix
]{revtex4-2}

\usepackage{graphicx}
\usepackage{dcolumn}
\usepackage{bm}
\usepackage{color}
\usepackage{xcolor}
\usepackage[normalem]{ulem}
\usepackage{hyperref}

\definecolor{darkgreen}{RGB}{0,130,0}
\newcommand{\adamo}[1]{#1}

\begin{document}

\preprint{APS/123-QED}

\title{AI sustains higher strategic tension than humans in chess}

\author{Adamo Cerioli\textsuperscript{1}, Edward D.~Lee\textsuperscript{2,3}, Vito D.~P.~Servedio\textsuperscript{2, *}}

\affiliation{
 \textsuperscript{1}Department of Mathematics, Physics and Computer Science, University of Parma, Parco Area delle Scienze, 7/A, 43124, Parma, Italy
}%
\affiliation{
 \textsuperscript{2}Complexity Science Hub, Metternichgasse 8, 1030, Vienna, Austria
}
\affiliation{
 \textsuperscript{3}Institute for Forest Ecology, BOKU, Peter-Jordan-Stra{\ss}e 82, 1190 Vienna, Austria
}%
\affiliation{
 \textsuperscript{*}Corresponding author: servedio@csh.ac.at
}%

\begin{abstract}
\noindent
Strategic decision-making requires balancing immediate opportunities against long-term objectives: a tension fundamental to competitive environments. We investigate this trade-off in chess by analyzing the dynamics of human and AI gameplay through a network-based metric that quantifies piece-to-piece interactions. Our analysis reveals that elite AI players sustain substantially higher levels of strategic tension for longer durations than top human grandmasters. We find that cumulative tension scales with algorithmic complexity in AI systems and increases linearly with skill level (Elo rating) in human play. Longer time controls are associated with higher tension in human games, reflecting the additional strategic complexity players can manage with more thinking time. The temporal profiles reveal contrasting approaches: highly competitive AI systems tolerate densely interconnected positions that balance offensive and defensive tactics over extended periods, while human players systematically limit tension and game complexity. These differences have broader implications for understanding how artificial and biological systems navigate complex strategic environments and for the deployment of AI in high-stakes competitive scenarios.
\end{abstract}


\maketitle

\noindent The aphorism that one may have won the battle but lost the war is encapsulated in the notion of a ``Pyrrhic victory.'' Costly short-term gains must be balanced against the longer-term uncertainties, opportunities, or challenges that may emerge in competitive environments. 
This presents a fundamental challenge in strategic decision-making for both biological and artificial agents across diverse competitive and cooperative landscapes \cite{vonNeumann1944, simon1955, bellman1957dynamic}, of which war \cite{weinsteinRebellionPolitics2007, bohorquezCommonEcology2009} and conflicts \cite{smithLogicAnimal1973, dedeoInductiveGame2010a, brushConflictsInterest2018} are poignant examples. More generally, organisms and individuals learn and adjust their behavior throughout their lifespan. This occurs during processes such as development (e.g., neural pruning \cite{woudeNoGains2019}), climate-dependent foraging \cite{piteraDailyForaging2018}, and the reorganization of social hierarchies \cite{magee20088}. Over longer evolutionary time scales, species continuously adapt to changing ecological niches \cite{lalandCulturalNiche2011}. In the context of an individual's lifetime, these can be characterized as strategic choices that involve an implicit trade-off between overfitting the local environment and maintaining the flexibility to adapt to unforeseen environmental conditions \cite{lee2024constructing}, in an echo of the speed-accuracy trade-off \cite{ratcliffModelingResponse1998}. In many real-world settings, complete measurements of the strategic environment, costs, and outcomes are not available. Chess, by contrast, is a game of perfect information, offering a well-studied, bounded setting.

As a result, chess has served as a model system for investigating strategy. It has well-defined rules, complexity, and rich history as a testbed for both human expertise and artificial intelligence (AI) \cite{shannon1950, simon1992chess, connors2011expertise, puddephatt2003chess}. Chess is a competitive game between two players, white and black, who take turns making moves with a single piece on a board with 8x8 squares as shown in Fig.~\ref{fig:network}. 
Both players start with 16 pieces, and the objective of the game is to capture the enemy's king. 
A key feature that makes the game interesting is the combinatorial explosion in potential moves, which prohibits exhaustive calculation. As a result, human players must make heuristic decisions about the possible long-term value of moves \cite{simon1992chess}. The nature of the trade-off is shaped by cognitive limitations, psychological pressures, and time constraints, often leading to biased or suboptimal decisions \cite{kiesel2009playing, lane2018chess, kahneman2011thinking}. 
In contrast, modern AI exhibits superhuman performance through two main strategies. One class involves deep reinforcement learning systems, such as AlphaZero and Leela Chess Zero, which develop strategies through self-play using deep convolutional neural networks. Alongside these, engines like Stockfish combine massive brute-force search trees with NNUE (Efficiently Updatable Neural Networks). While Stockfish originally relied on handcrafted evaluation functions, modern versions utilize machine learning to train their evaluation layers, blending classical search efficiency with neural network-based intuition.
These differences suggest that the trade-offs AI engines make are distinct from those made by human players. Still, there is little characterization of the properties that distinguish gameplay.

Here, we compare humans to AI play by analyzing the network structure of piece-to-piece interactions \cite{farren2013analysis} over many chess games. We consider thousands of games across a broad range of scenarios, including games with players of all skill levels and official AI tournaments between competitive chess engines, such as Leela Chess Zero and Stockfish, held between 2019 and 2024. In these tournaments, the engines operate adaptively, allocating varying amounts of time to different moves. 
In addition to these tournament games, we include simulations in which we run Stockfish at fixed search depths. This provides a controlled baseline representing a spectrum of computational constraints and playing strengths, allowing us to observe how strategic tension scales with algorithmic complexity.

We find that the complexity of the game, as measured by ``strategic tension,'' follows, on average, a regular curve, whose features correspond to the different stages of the game. 
The curve is characteristically different for human players compared with AI players. Our observations suggest that humans and AI exhibit distinct behaviors in complex strategic environments, raising questions about the consequences of delegating competitive real-life scenarios to AI.

\section{Defining Tension in Chess}
\noindent
In chess, tension refers to a state of unresolved tactical interaction where multiple pieces from both sides occupy squares from which they can capture or be captured by one another. Unlike a direct exchange of pieces, which resolves the interaction and simplifies the board state, tension represents a ``latent'' complexity. It persists as long as players maintain these mutual threats without acting upon them, forcing both competitors to constantly evaluate the shifting consequences of a potential explosion of captures. By formalizing these threats into a network of offensive and defensive relationships, we can quantify this tension as the density and structure of piece-to-piece interactions.

\begin{figure}[!htbp]
\centering
\includegraphics[width=1\linewidth]{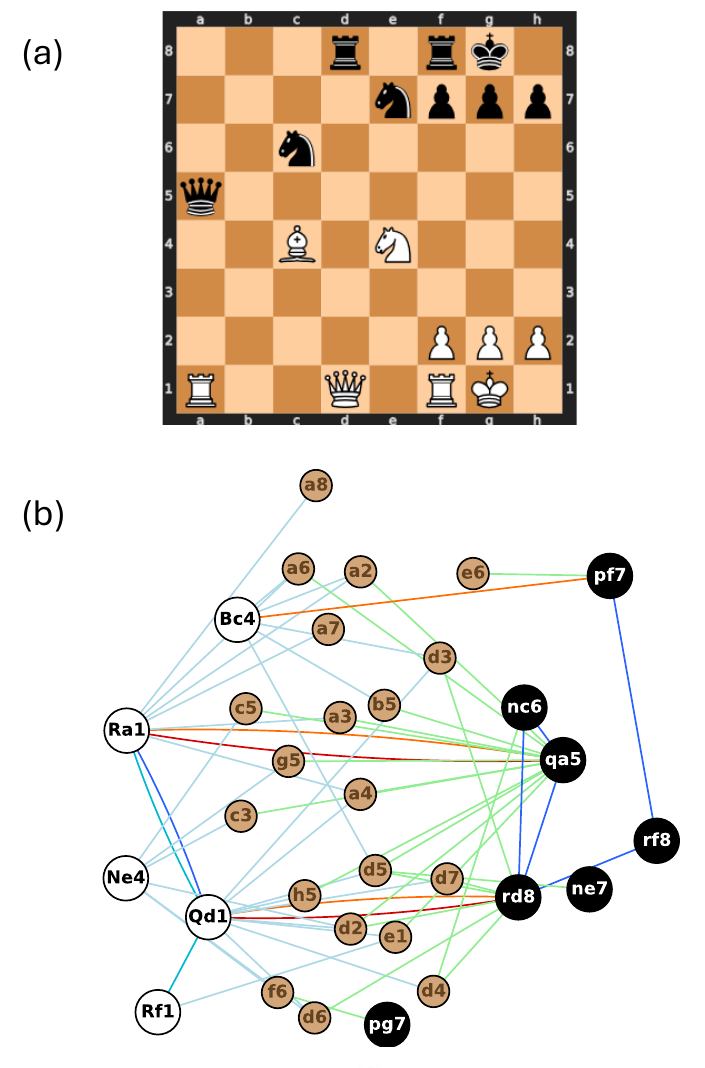}
\caption{\label{fig:network} One example of tension network. (a) Chessboard at the ply 44 during a game between the two grandmasters, Magnus Carlsen (white) and Alexey Sarana (black), in the Titled Tuesday tournament, Round 6, held on chess.com on November 26, 2024. (b) The network represents the interactions on the chessboard. The chess pieces are depicted as dark or light circles, depending on their color. The attack links are shown in red when originating from black and in orange when originating from white. Defense links are turquoise when white is defending and blue otherwise. Control links are light blue (originating white) and light green (originating black). Small, brownish nodes represent vacant squares under control. The chosen ply corresponds to the highest tension reached during the game, as determined by the maximum eigenvalue of the adjacency matrix. 
}
\end{figure}

Given the state of a chessboard as in Fig.~\ref{fig:network}a, we consider the set of offensive and defensive relationships at each ply (half-move). An attacking piece can capture an opposing piece in a single move. The defending piece would threaten any piece that would capture the piece it is defending. This is the corresponding network in Fig.~\ref{fig:network}b, where a node is either a chess piece $p_i$ or a vacant square $s_i$, and the links satisfy combinations of the following three conditions:
\begin{description}
    \item[Attack link]  We define an attack link from $p_i$ to $p_j$ if $p_i$, a piece from one color, can legally capture $p_j$, a piece from the opposing color. Attack links reflect direct threats, highlighting where immediate exchanges could occur. 
    \item[Defense link] We draw a defense link from $p_i$ to $p_j$ if there exists an opponent's piece $p_k$ that can legally capture $p_j$ and after $p_k$'s move, $p_i$ would be able to legally capture $p_k$ in a single move. Kings can defend other pieces, but cannot be defended under this definition. This distinction arises because the King is never captured in chess; hence, the legal mechanism of a reciprocal capture (or ``re-capture'') cannot apply to it. Defense links reveal the layers of protection that determine whether an attack can be carried out safely or must be postponed. 
    \item[Control link] We define a control link from $p_i$ to a vacant square $s_j$ if there exists an opponent's piece $p_k$ that can legally move to $s_j$ and after $p_k$'s move, $p_i$ would be able to legally capture $p_k$ on $s_j$ in a single move. Control links, finally, represent subtler forms of influence, where empty squares become contested zones. 
\end{description}
In constructing the network, we disregard which side moves, focusing instead on the full set of interactions at each ply to capture the board's full state. For the main part of the analysis, we consider only the connectivity of the network and do not distinguish the links from one another; we indicate where we consider link types in the analysis that follows.
By construction, the network has at most 64 connected nodes, each corresponding to either a chess piece or a vacant square.

Among the many ways to characterize the interaction network, we focus on how game complexity evolves, reflected in features such as the number of interacting pieces and loops \footnote{A loop consists of a closed sequence of interactions between pieces forming a cycle within the network (e.g., a white rook on b4 defends a white knight on e4; the knight, in turn, attacks a black bishop on d6, which attacks the rook on b4).}. We interpret this complexity both in terms of structural richness and strategic fragility, namely the potential for cascading consequences following a suboptimal move. In dense, highly connected configurations, even a small error can propagate across the network, substantially shifting the advantage. To capture this latent vulnerability, we introduce an original definition of \textit{strategic tension}, quantified via the maximum eigenvalue of the adjacency matrix, $\lambda_1$. While previous and similar tasks in the context of chess have adopted measures based on betweenness centrality~\cite{barthelemy2025fragility}, our definition focuses entirely on the network's spectral properties.

The maximum eigenvalue, or spectral radius, reflects network functionality, encompassing both its structural features and dynamical properties~[\onlinecite{pillai2005perron}, \onlinecite{castellano2017relating}]. This can be understood via topological entropy, which captures the complexity and unpredictability of walks in a network~[\onlinecite{adler1965topological}]: the spectral radius is proportional to the asymptotic exponential growth rate of walks in the network. Higher $\lambda_1$ indicates greater dynamical complexity, consistent with tension as a measure of uncertainty or potential instability. Importantly, $\lambda_1$ is bounded between the average and maximum node degree, $\langle s \rangle \le \lambda_1 \le s_{\max}$, linking global connectivity with local heterogeneity~[\onlinecite{van2023graph}]. 
Thus, the spectral radius serves as a natural metric for evaluating tension at any given state of the board. It acts as a bridge between the game's structural properties, underpinned by the Perron–Frobenius theorem, and its dynamical behavior, as captured by entropy-like measures (see Appendix~\ref{subsec:topological entropy}). The undirected nature of the network treats interactions as reciprocal. This choice is not merely a simplification but a requirement for numerical stability and theoretical consistency. In a directed representation, tactical configurations are often acyclic, which causes the spectral radius to vanish and fails to reflect the board's actual complexity.

While we do not distinguish between link types for strategic tension \footnote{Although we treat all links equally for the main macroscopic metric, we also performed a structural deconstruction of the network. As we will show, isolating specific subsets of links (e.g., generating attack-only or attack-and-defense networks) yields qualitatively similar tension trajectories.}, the resulting network captures the global interplay among pieces and squares that creates and sustains tension on the chessboard. Building on this global metric, we also evaluate the \textit{tension load}, defined as the cumulative sum of the strategic tension up to ply 150, which captures the total tactical effort sustained throughout the game. To investigate specific micro-structural dynamics, however, we explicitly leverage the differentiation between link types to analyze the \textit{attack-defense balance}, an observable that quantifies the difference between offensive and defensive interactions on the board. Here, attack and defense links can have weight 1 or 2, depending on whether the corresponding conditions are satisfied by one or both of the pieces involved. Control links always have weight 1.
Furthermore, to quantify the structural heterogeneity of the network, we analyze the \textit{degree variation}, defined as the ratio of the standard deviation of the node degrees to their mean ($\sigma_k / \mu_k$). This observable captures how unevenly the tactical interactions are distributed among the pieces on the board.

\section*{RESULT 1: humans peak, AI sustains STRATEGIC TENSION}
\begin{figure}[t]
\centering
\includegraphics[width=1\linewidth]{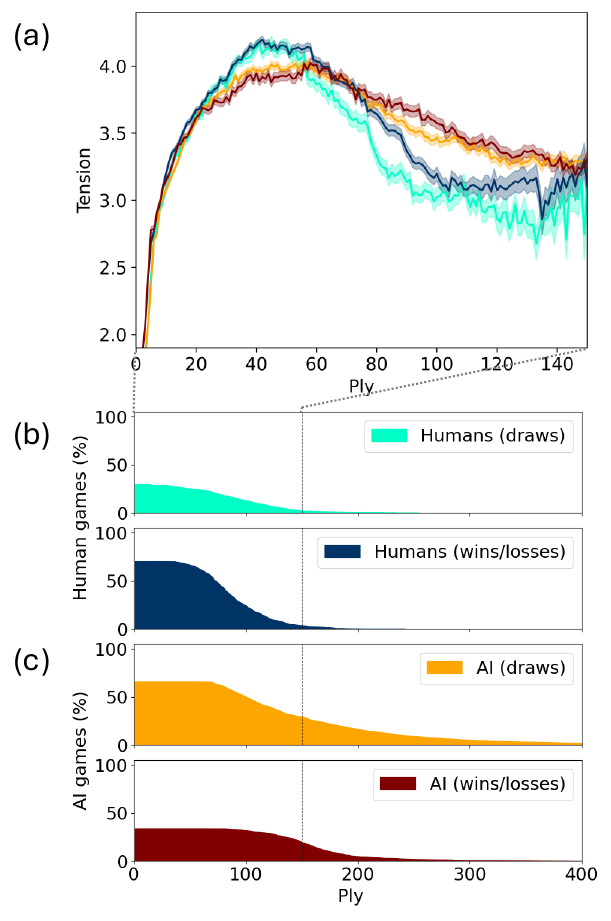}
\caption{\label{fig:tension} Tension values during games for both humans (grandmasters) and AI (Stockfish and Leela Chess Zero). We consider 1,200 games for each, with human games obtained from \texttt{https://www.pgnmentor.com} and AI games from \texttt{https://tcec-chess.com}, which hosts official engine-versus-engine tournaments (a) Average tension is represented in light blue and blue for humans, and in orange and red for AI, respectively. For both groups, we distinguish between games that end in a draw and those that result in a win/loss, represented with light and dark shades, respectively. (b) Survival curves of human games, showing the percentage of total games that remain active at each ply, categorized by their final outcome (draw vs. decisive win/loss). The steady decline illustrates the statistical distribution of game lengths. (c) We conducted the same survival analysis for AI games, highlighting differences in game duration and outcome patterns.}
\end{figure}

\noindent 
We illustrate the typical evolution of the strategic tension across multiple games in Fig.~\ref{fig:tension}. We can distinguish four main phases. The first phase, roughly corresponding to the opening stage up to around ply 30, features a buildup in which pieces approach each other and contested squares come under threat, leading to an abrupt increase in tension. The second phase, spanning approximately from ply 30 to ply 60, represents peak complexity, where the board remains highly interconnected, and the tension stays elevated. The third phase exhibits a gradual simplification of the tension network, as piece exchange and loss steadily lower the overall tension. Finally, the games reach a relatively stable value of tension, but at this point available statistics become increasingly sparse, especially for games played between humans that typically end earlier
(Fig.~\ref{fig:tension}b and \ref{fig:tension}c). These patterns remain robust across control benchmarks, demonstrating that the observed dynamics strongly reflect deliberate strategic choices. As detailed in Appendix~\ref{subsec:random_benchmark}, our findings hold against null models of random moves and varying time controls: \adamo{notably, while pure random play exhibits extremely high tension, introducing even minimal strategic intent causes a sharp collapse, which then gradually recovers only as the playing strength increases.} We further contrast a real competitive game with an artificial, tension-maximizing scenario, alongside a network ablation study (see Appendices~\ref{subsec:example_tension}, \ref{subsec:max_tension}, and \ref{subsec:Ablation_Study_Tension}). Thus, the dynamics of tension capture the overall contours of the game, including the opening buildup, the combinatorially complex midgame, and the relaxation leading up to the endgame as discussed in the theory of chess play \cite{de2014thought}. 

Moreover, despite the overall similarities in the typical trajectory of strategic tension between human and AI play, there are notable differences. Maximum tension is higher in human-played games than in engine-played games. Up to approximately ply 40, human players typically follow established opening repertoires from theoretical databases, which often lead to well-studied positions with numerous direct piece interactions, including attacks, defenses, and immediate threats \cite{DeMarzo2023Quantifying, Campbell1999Knowledge}. These theory-driven positions generate the highest points of tension as players transition to original play in the midgame (see Appendix~\ref{subsec:tension_peak}). 
Once players leave the realm of memorized theory and must rely on their own calculations and evaluations, they tend to simplify positions. Human players reduce tension more rapidly and, as we discuss below, resolve piece conflicts sooner, often favoring attack-oriented configurations. In contrast, state-of-the-art engines tend to maintain more distributed positional advantages that unfold gradually over multiple moves, leading to slower tension decay.

A secondary difference emerges in how strategic tension declines in drawn vs.\ decisive games. Drawn games between humans show a rapid, substantial drop in tension compared to decisive games. This agrees with the faster decrease in the number of pieces for drawn games between human players (see Appendix~\ref{subsec:additional_properties}). AI games show a contrasting pattern. Drawn games show significantly more (not fewer) pieces on the board relative to decisive games in the endgame, and tension does not differentiate strongly between game outcomes.

\section{Result 2: Game outcomes diverge before peak strategic tension}
\noindent
To relate the structural complexity of the interaction network to game outcomes, we compare the evolution of strategic tension with the algorithmic evaluation provided by Stockfish (see Appendix~\ref{subsec:stockfish_evaluation}). Inspecting the evaluation score, we identify \textit{bifurcation points} in game trajectories, typically around ply 30 (move 15), where decisive and drawn games begin to diverge for human players during the opening-to-midgame transition. A similar behavior is observed in AI games, where a local maximum or inflection point in the evaluation profile precedes the divergence of outcomes. Notably, in all games, this divergence occurs \textit{before} strategic tension reaches its absolute peak. In fact, we find that this ordering happens more consistently than this comparison of averages would suggest. By aligning games by the inflection point (further detailed in Appendix~\ref{subsec:stockfish_evaluation}), we show that tension peaks immediately following the decisive evaluation divergence before undergoing a sustained decline.
This ordering suggests that the game’s outcome is largely determined during the \textit{ascent} toward peak tension, before the subsequent reduction in tension differentiates decisive games from draws. In chess terms, foundational advantages, such as positional pressure, space, or superior piece coordination, are established as the board becomes increasingly complex.

The ensuing reduction in tension reflects a phase of \textit{conversion} or \textit{simplification} from the tension peak, often marking the resolution of a struggle in which one side has attempted to generate complications. Consistent with established principles of high-level play, players with a decisive advantage tend to trade pieces to limit counterplay \cite{silman2010reassess}. Our results confirm that this reduction is not the cause of the outcome, but rather its pursuant: both humans and engines simplify positions once a win or a forced draw is already favored.

\section*{RESULT 3: Variability in strategic tension reflects different gameplay}
\noindent
While tension seemingly captures only general and coarse features of the game, it is correlated with key aspects of gameplay that distinguish humans from AI (see Appendix~\ref{subsec:additional_properties}).

\textit{Number of links and loops:} AI interaction networks exhibit more links and loops, indicating a richer web of mutual threats, defenses, and controls. However, the maximum average number of loops is higher in human games, reflecting the peak of human tension.

\textit{Number of pieces:} AI maintains a higher number of pieces over the trajectory of games on average. This reflects a weaker tendency to simplify the game by removing pieces from the board. From midgame to endgame, the small number of pieces limits the strategic tension typically found in human games.

\begin{figure}[t]
\centering
\includegraphics[width=1\linewidth]{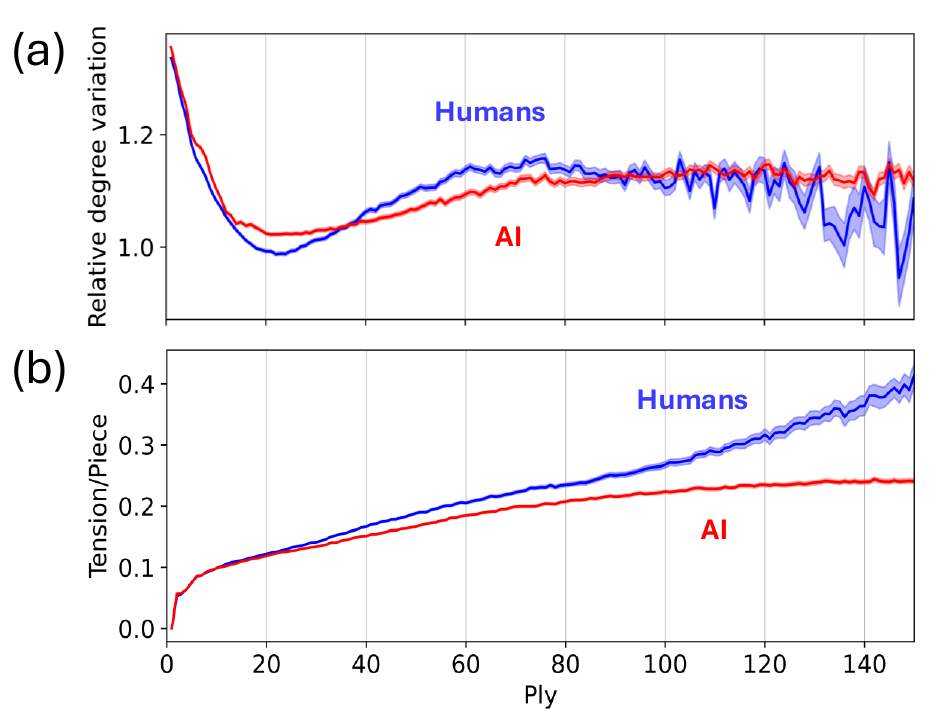}
\caption{\label{fig:properties} Comparison of structural properties across human and AI games. We analyze the set of games shown in Fig.~\ref{fig:tension}. Two measures are considered: the average standard deviation of the degrees associated with chess pieces, normalized by the mean of the degrees, and the average tension per piece. Red is for AI and blue for human games.}
\end{figure}

\textit{Degree variation:} The noise coefficient of the degree distribution over chess pieces mirrors the rise and fall of strategic tension, first falling then rising. 
While human and AI games display a broadly similar global trend, human play exhibits more pronounced variation as shown in Fig.~\ref{fig:properties}a. In particular, during the opening phase, the relative degree variation decreases markedly, reflecting a positioning phase in which interactions are more evenly distributed across pieces as tension accumulates. As the game progresses, however, the interaction network in human games exhibits greater variability, eventually returning to levels similar to those observed in AI games, or even lower.

\textit{Tension per piece:} As shown in Fig.~\ref{fig:properties}b, the tension per piece $\lambda_{1}(t)/n_t$ remains consistently higher across all plies in human games. While human play shows approximately linear increases in tension per piece throughout the game, AI games display sublinear progression that appears to plateau around ply 80. 
Human games involve more frequent captures of pieces (see Appendix~\ref{subsec:additional_properties}), but the remaining pieces are then involved in a relatively greater number and greater complexity of interactions. Note that the tension (Fig.~\ref{fig:tension}) and the number of pieces are not simply related. Crowding constrains the formation of interaction links as described further in Appendix~\ref{subsec:tension_number_pieces}.

\begin{figure}[t]
\centering
\includegraphics[width=1\linewidth]{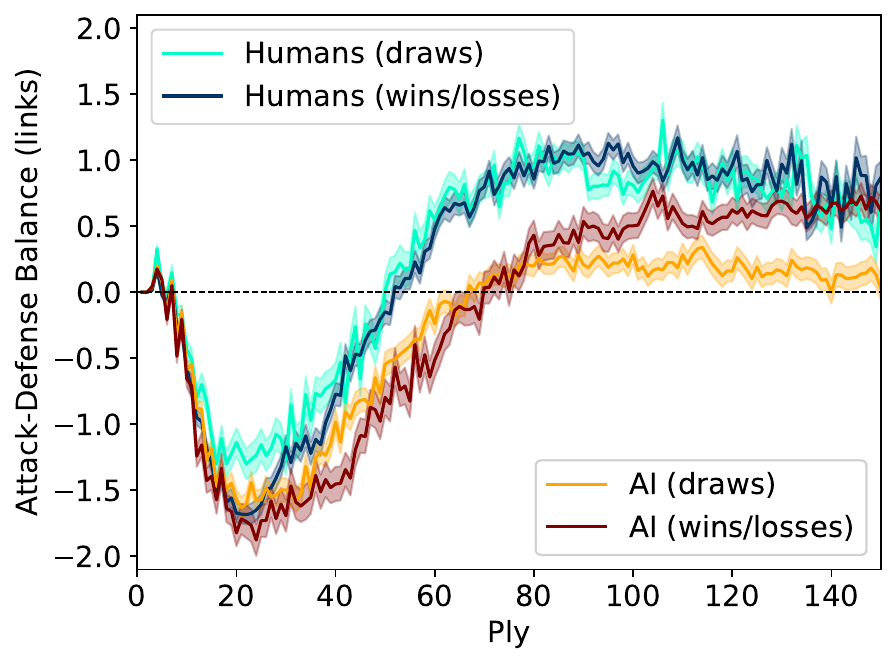}
\caption{\label{fig:attacks_defenses} Balance between attack and defense links across human and AI games. We analyze the set of games shown in Fig.~\ref{fig:tension}. For both human and AI games, we distinguish between games that end in a draw and those that result in a win/loss, represented with light and dark shades, respectively.}
\end{figure}

\textit{Attack-defense balance:} When distinguishing between attack and defense links, we find that the surplus of attack links over defense links, as shown in Fig.~\ref{fig:attacks_defenses}, moves inversely with the tension profile. It initially decreases during the early, more defense-oriented phase of the game. As the game progresses, attack links gradually outnumber defense links. The curve is shifted to more positive values in human games than in AI games. 

Overall, we find that strategic tension aligns with basic gameplay metrics, such as the number of pieces and, in particular, the presence of loops in the interaction graph. At the same time, it is informative to relate it to the balance across pieces on the board and to how pieces are arranged in formations. Taken together, our results indicate that AI games retain more pieces, exhibit a more balanced network of attack–defense connections, and display less variability in the degree distribution across pieces. 

\section*{RESULT 4: Cumulative TENSION SCALES WITH SKILL AND COMPUTATIONAL COMPLEXITY}

\begin{figure}[t]
\centering
\includegraphics[width=1\linewidth]{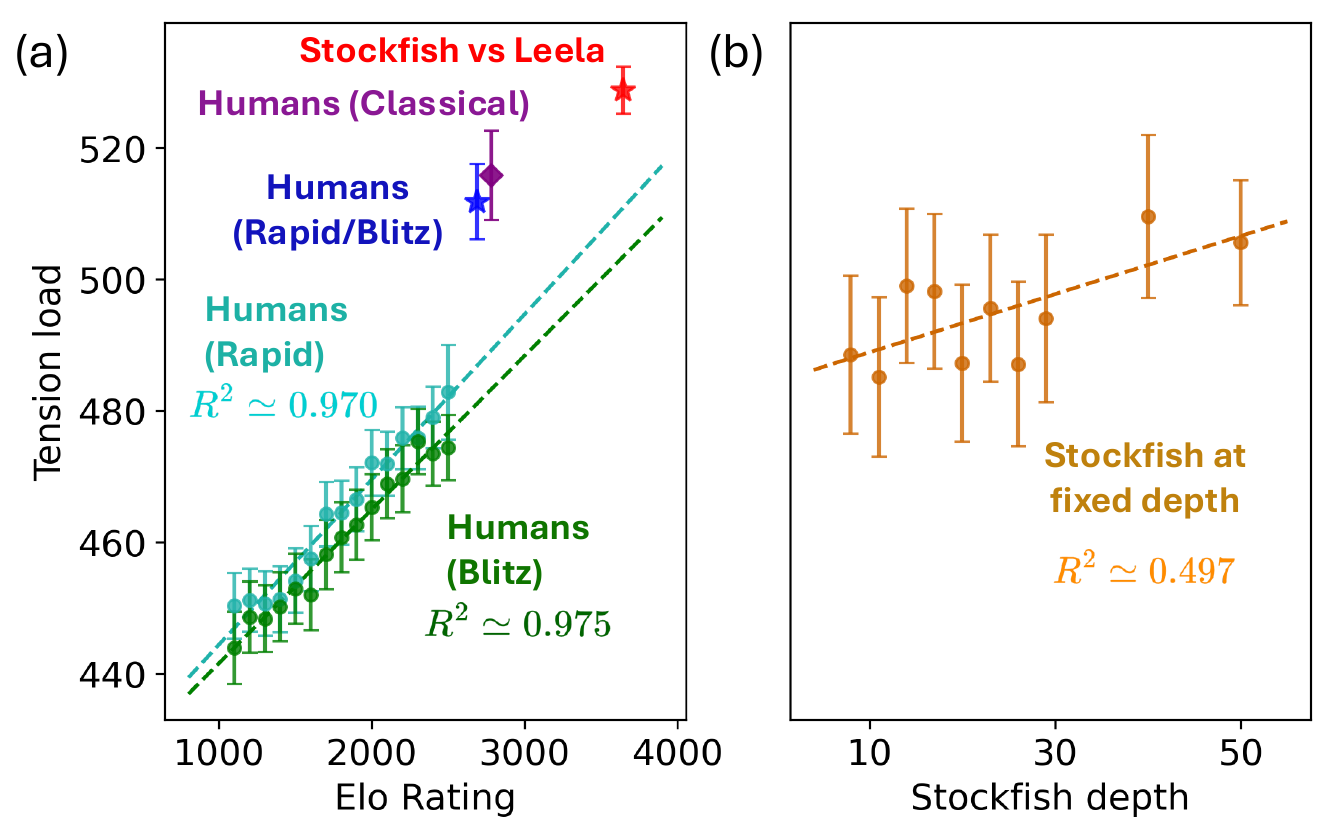}
\caption{\label{fig:tension_load} Tension loads for different Elo ratings and Stockfish depths. (a) Tension loads are shown for rapid games (turquoise dots) and blitz games (green dots). The plot also includes the peaks of top human players in rapid/blitz (blue star) and classical games (purple diamond), as well as elite AI matches between Stockfish and Leela from official TCEC tournaments (red star). All data points are plotted against their estimated Elo ratings. Each dot/star represents the average across 1,200 games. The blue and red stars were derived from the data analyzed in Fig.~\ref{fig:tension}. For reference, blitz games are played with very short time controls (typically 3–5 minutes per player), rapid games with intermediate time controls (around 10–25 minutes per player), and classical games with long time controls (often 60 minutes or more per player), allowing deeper strategic play. (b) We represent tension loads from games between versions of Stockfish at fixed depth levels (orange dots), computed from a smaller sample of 120 games per dot. The two panels also show the corresponding linear regressions (Rapid: $R^2=0.970$, $p=2.55 \times 10^{-11}$; Blitz: $R^2=0.975$, $p=8.79 \times 10^{-12}$; AI: $R^2=0.497$, $p=0.023$).}
\end{figure}

\noindent While engines can maintain higher tension for a longer duration compared to humans, does this difference depend on skill level? We consider the tension load across different Elo ratings for human players in Fig.~\ref{fig:tension_load}a \footnote{In principle, Stockfish has a search depth parameter that serves as a proxy for skill level. Leela has a time limit parameter during which it is permitted to assess the board state and make a decision. At a few dozen search depths, the tension load is limited to about 450, similarly for Leela at a few seconds. Since the games that are shown go far beyond these limited parameter values with supercomputers, this suggests that algorithmic complexity increases the ability to handle tension.}. 
Load changes substantially with Elo, showing an approximately linear dependence in human games. Moreover, across all Elo values, rapid games consistently exhibit slightly higher tension loads than blitz games. The gain in load is most pronounced in classical games, where players have significantly more time to make decisions. 
Particularly notable are top-tier grandmasters who show the highest amounts of tension load in all types of games, but especially under classical time controls. Elite classical play is often characterized by maintaining a complex struggle, where grandmasters sustain tense, fluid positions that test the opponent’s stamina and depth of calculation over many hours \cite{Kotov1971Think, Nunn1997Secrets, degroot1965thought}. Time-constrained players experience larger performance drops on complex decisions and therefore tend to choose safer, less strategically risky continuations, a noted concern in the chess community~\cite{Calderwood1988Time, Carow2025Time}. 
These observations all point to the conclusion that longer time controls allow human players to create and manage more complex positions \cite{Calderwood1988Time}.

For comparison, we examine how the tension load changes for Stockfish, where we control the computational complexity of the algorithm by adjusting the search depth at a fixed value for simulated games in Fig.~\ref{fig:tension_load}b. Up to search depths of 50, or incredibly adept players, the simulation results indicate that increasing depth leads to higher tension. As a reference, previous experiments using a different engine (Houdini) reported a roughly linear relationship between Elo rating and search depth up to 20 plies~[\onlinecite{ferreira2013impact}]. 
Overall, this indicates that increased tension load reflects greater computational demands on the player. 

\adamo{To verify whether this scaling reflects universal strategic choices or algorithmic bias, we provide a cross-population validation in Appendix~\ref{subsec:Skill-Calibrated_Comparison}. Mapping both populations onto a shared Elo-calibrated axis reveals that the tension-strength scaling is robust, although synthetic agents maintain a consistently distinct tactical posture compared to humans of equivalent skill. Overall, this confirms that the coupling between tension and computational depth is a fundamental property.}

\section{Discussion}
\noindent
We investigate the dynamics of strategic tension in chess using an interaction network that reflects board layout. The central hypothesis is that distinct tension profiles would emerge between skill-matched human and AI play, reflecting differing strategic approaches. Our observations confirm this hypothesis and reveal that state-of-the-art chess engines (Stockfish at high search depths and Leela Chess Zero at advanced neural network settings) consistently maintain significantly higher levels of strategic tension for longer durations than even the best human grandmasters. 

Essential to the test of the hypothesis is the focus on players with commensurate skill levels since the overall complexity of a game is bounded by the weaker player (inspired by recent methodologies~[\onlinecite{peri2026smart}], we can show that a wider skill gap in simulated matches reduces the measured tension in Appendix~\ref{subsec:depth_gap}). Another limitation is that tension does not clearly distinguish between strategies that are simply stronger and those that are fundamentally artificial. The test with Stockfish at varying search depths supports the central finding that tension grows with algorithmic complexity, both in AI and in human play. \adamo{Notably, this growth requires intentionality, emerging only after a minimum degree of strategic intent is introduced to separate purposeful play from purely random benchmarks (Appendix~\ref{subsec:random_benchmark}). In game theory, the macroscopic divergence between purely random actors and those guided by even a minimal payoff matrix is well-documented; the injection of tiny strategic incentives is known to trigger non-linear phase transitions that collapse the active degrees of freedom of the system's state space~[\onlinecite{savit1999adaptive}].}

We speculate that the sustained tension in the AI play reflects the deferral of immediate opportunities and facilitates the emergence of latent strategic options that materialize only in later phases of the game. This kind of strategic decision-making would be facilitated by the ability to handle greater computational complexity.
Correspondingly, we find that human players exhibit a linear relationship between cumulative tension and skill level (Elo rating). Similarly, Stockfish's tension load increases with computational complexity, as reflected in its search depth. 
Far beyond these examples are AI engines running on supercomputers. Superhuman computational capacity obviates the need to simplify upon leaving theoretical territory; AI engines do not need to rely on heuristics and intuition, such as familiarity with positions in opening theory that heavily guide expert human play. These observations clarify why human players experience a tension drop: they can navigate extreme complexity primarily while relying on the extensive memorization of deep theoretical lines \cite{degroot1965thought, charness1992expert}, often being forced to simplify once ``out of book". In contrast, AI dynamically sustains high strategic tension throughout the game, driven by superior objective calculation.

The increase in tension with computational load is likely due to different factors in the two groups, reflecting their nature as artificial versus biological systems. In AI, tension rises as engines gain computational capacity and play at higher levels, where controlling more pieces and squares maximizes options and exploits opponents’ mistakes \cite{berliner1999system}. In humans, more skilled players can handle complex positions and maintain higher‑tension situations without simplifying \cite{Kuchelmann2022Expertise, connors2011expertise, Chassy2011Measuring, Gobet2016Understanding}. Indeed, humans may tend to simplify under certain conditions, possibly due to limitations in working memory, attention, or susceptibility to psychological pressure \cite{kahneman2011thinking, lazarus2000emotions}. This interpretation of the results could also suggest that human players may sometimes actively steer games toward drawn resolutions, reflecting deliberate risk minimization and metagame considerations typically absent in AI play. The simplification of high-tension positions thus serves not only strategic purposes (such as pruning the calculation tree into a strict set of candidate moves \cite{Kotov1971Think}) but also as an adaptive response to the computational demands of accurate tactical calculation, perhaps allowing players to sustain the level of accuracy needed for strong play within cognitive limits \cite{delafuente2011flexible}.
Of course, such limits come from the evolutionary trade-offs reflected in the brain, where short-term costs, such as losing a chess game, are outweighed by long-term benefits in metabolic efficiency and general, rather than domain-specific, intelligence \cite{snell-roodBrainSizeGlobal2009, burgerAllometryBrainSize2019}.

Several avenues for future research emerge from this work. 
Extending this methodology to other complex strategic games, such as Go \cite{silver2017mastering}, Shogi, or even real-time strategy e-sports, could test the universality of the observed dynamics across different game structures (e.g., imperfect information and varying spatial dynamics).
Moreover, the AI engine used (Stockfish) represents a single architecture; other engines with different algorithms, such as pure neural network-based engines like AlphaZero, may exhibit different tension dynamics. While we focused on Stockfish and Leela due to their prominence and availability, our analytical framework and open-source code can readily be applied to other UCI-compatible engines (Universal Chess Interface), enabling broader comparative studies. With a more direct focus on tension, directly investigating the cognitive and physiological correlates of our tension metric in human players, for example, by combining game analysis with eye-tracking, think-aloud protocols, or stress indicators such as galvanic skin response, could help elucidate the underlying mechanisms driving changes in tension.

The observed differences between artificial and human systems could have implications for AI designed to interact with humans, particularly as AI is expected to play an increasing role in teaching and training across many domains. In chess, a notable step in this direction is the Maia engine \cite{anderson2020maia}, which is explicitly designed to model human play rather than to optimize engine moves. Moreover, while direct extrapolation requires caution (after all, real-world conflicts involve imperfect information and stochastic shocks), examining how tension is sustained in other complex adaptive systems, such as economic markets or geopolitical negotiations, can provide insight \cite{schelling1966arms}. %
The finding that increased computational complexity leads AI to sustain tension raises questions for stability in international relations \cite{simmons-edlerPositionAIPowered2024}, where the use of AI potentially enables more complex standoffs but also increases the stakes of miscalculation \cite{jervis1978cooperation, guttieriIntegrativeComplexity1995, sakuwa2019origins}, creating an AI-enabled parallel to \textit{Dr.~Strangelove}.

\section{Methods}

\subsection{Data}
\noindent
Chess games are commonly stored in the PGN (Portable Game Notation) format. This standard plain-text representation records the sequence of moves in a game, along with optional metadata such as player names, event, result, and time controls. This format is human-readable and machine-parsable, making it ideal for large-scale game analysis.

The games between top-level human players across Rapid, Blitz, and Classical formats were obtained from \texttt{https://www.pgnmentor.com}, which provides thousands of PGN-formatted games organized by player or sorted by opening. This source includes a comprehensive archive of historical and modern games, making it well-suited for comparative analysis.

Competitive matches between Stockfish and Leela Chess Zero were collected from the \texttt{https://tcec-chess.com/} website, which regularly publishes high-level engine-versus-engine games in PGN format. These games represent the cutting edge of chess engine performance under standardized tournament conditions. The typical time control is 120 minutes per player, with a 12-second increment added after each move.

These datasets served as the basis for the analyses presented in Fig.~\ref{fig:tension}.

In addition, a large set of rapid and blitz games played by human players with Elo ratings ranging from 1000 to 2700 was downloaded from \texttt{https://lichess.org/}, an online platform that provides a vast archive of games. This dataset broadly represents amateur to expert players and is ideal for analysis because games are subject to anti-cheating controls, ensuring reliable data across a broad skill spectrum.

Together with the previously described datasets, these data formed the basis for the analyses presented in Fig.~\ref{fig:tension_load}.

\subsection{Software}
\noindent
All operations related to chess move generation, rule validation, and position analysis were carried out using the python-chess library (v1.9.4). This comprehensive library offers tools for representing chess boards, parsing PGN files, and computing legal moves according to FIDE rules. Furthermore, it provides native support for communicating with UCI-compatible engines such as Stockfish, which allowed us to automate position evaluation and simulate engine-vs-engine games at varying search depths.

\subsection{Generating Stockfish games at varying depths}
\noindent
To generate games between Stockfish and itself at various search depths, we followed a procedure that balanced realism and controlled variability. While there is no universally accepted mapping between Stockfish's search depth and Elo rating, prior work ~[\onlinecite{ferreira2013impact}] has estimated approximate correspondences. Based on these estimates, we selected a depth range roughly aligned with the playing strength of top human players and beyond.

To introduce diversity in the opening phase and avoid deterministic repetition, each game was initialized using the first 12 plies (typically covering the opening) from randomly selected games played by top-level human players. After this opening phase, the rest of the game was played entirely by two instances of Stockfish, one playing each move at either depth $D$ or $D+1$, chosen randomly, and the other at either depth $D$ or $D-1$, also chosen randomly. This asymmetry introduced slight variability in evaluation and decision-making while preserving a consistent overall depth level $D$.

\section*{Code availability}
\noindent
Python code used to compute tension and generate the Stockfish self-play games is available on GitHub at \href{https://github.com/Draghellon/Tension\_Chess}{https://github.com/Draghellon/Tension\_Chess}.

\section*{Authors' contributions}
\noindent
AC led the definition of the tension metric, with contributions from the other authors, and carried out the code development, numerical analyses, and figure generation.
EDL performed simulations on a computing cluster using Stockfish at high depths.
VDPS provided the original inspiration for the project, supervised its development at all stages, and independently cross-checked the results.
All authors participated in the discussions and contributed to writing the manuscript.

\section*{Acknowledgements}
\noindent
We thank V.~Latora for valuable discussions, particularly regarding the inclusion of Fig.~\ref{fig:attacks_defenses}, and M.~Barthelemy for useful discussions and for sharing his code, which we used as a baseline for our analysis. 
This work was partly funded by the City of Vienna, Municipal Department 7, and the Federal Ministry of the Republic of Austria for Climate Action, Environment, Energy, Mobility, Innovation, and Technology as part of the project GZ 2023-0.841.266. It was also partially funded by the Austrian Science Fund (FWF) 10.55776/ESP127.

\appendix
\section{Why the spectral radius as a measure of tension}
\label{subsec:topological entropy}
\noindent
Topological entropy, $h_\mathrm{top}$, is a concept from dynamical systems theory that quantifies the exponential growth rate of the number of distinguishable orbits or, in the context of graphs, the diversity of walks. Topological entropy was first introduced in 1965 by Adler, Konheim, and McAndrew~\cite{adler1965topological} as a measure of the complexity of a dynamical system. 

Building upon this foundation, Bowen~\cite{bowen1971entropy} introduced a metric-based formulation of entropy. Instead of relying on open covers, Bowen defined it through the asymptotic growth rate of the maximum number of distinguishable finite sequences (or trajectories) of the system's states at a given resolution. Following this trajectory-based perspective, Dinaburg~\cite{dinaburg1970correlation} made a crucial theoretical breakthrough by rigorously demonstrating the correlation between topological and metric entropies.

Here, we report a few calculations that illustrate this interpretation of the spectral radius. Given a network and its corresponding adjacency matrix $A$, the number of walks of length $n$ is given by
\[
N(n) = \sum_{i,j} (A^n)_{ij}.
\]

This expression corresponds to the sum of all entries of the matrix $A^n$, which can be decomposed using its eigenvalues and eigenvectors:
\[
A^n = \sum_{i=1}^{m} \lambda_i^n \mathbf{v}_i \mathbf{u}_i^T
\]
where $\lambda_i$ is the $i$-th eigenvalue, $\mathbf{v}_i$ is the corresponding right column eigenvector, and $\mathbf{u}_i^T$ is the left row eigenvector.

When modeling an undirected network, the adjacency matrix $A$ is symmetric ($A = A^T$). According to the Spectral Theorem, the left and right eigenvectors of a real symmetric matrix are identical ($\mathbf{u}_i = \mathbf{v}_i$) and form an orthogonal basis. Therefore, the spectral decomposition simplifies to:
\[
A^n = \sum_{i=1}^{m} \lambda_i^n \mathbf{v}_i \mathbf{v}_i^T
\]

To calculate the total number of walks $N(n)$, we need to sum all entries of $A^n$. In matrix notation, the sum of all elements of a matrix can be elegantly expressed by pre- and post-multiplying it by the all-ones vector $\mathbf{1} = (1, 1, \dots, 1)^T$:
\[
N(n) = \mathbf{1}^T A^n \mathbf{1}
\]
By substituting the spectral decomposition of $A^n$ for a symmetric matrix into this equation, we obtain:
\[
N(n) = \mathbf{1}^T \left( \sum_{i=1}^{m} \lambda_i^n \mathbf{v}_i \mathbf{v}_i^T \right) \mathbf{1} 
\]
Because matrix multiplication is distributive and associative, we can rearrange the terms:
\[
N(n) = \sum_{i=1}^{m} \lambda_i^n (\mathbf{1}^T \mathbf{v}_i) (\mathbf{v}_i^T \mathbf{1})
\]
Notice that the term $\mathbf{1}^T \mathbf{v}_i$ is a scalar representing the sum of all components of the $i$-th eigenvector. Let us define this sum as $S_i = \sum_{j=1}^{m} (\mathbf{v}_i)_j$. Since the dot product is commutative, $\mathbf{1}^T \mathbf{v}_i = \mathbf{v}_i^T \mathbf{1} = S_i$. Therefore, the equation reduces to:
\[
N(n) = \sum_{i=1}^{m} S_i^2 \lambda_i^n
\]
This final expression demonstrates that the total number of walks of length $n$ is a linear combination of the network's eigenvalues raised to the power of $n$, weighted by $S_i^2$, the squared sum of the elements of their corresponding eigenvectors.
To evaluate the exponential growth rate of the number of possible paths in the infinite limit ($n \to \infty$), we calculate the topological entropy of the network, denoted as $h_\mathrm{top}$. It is defined as:
\[
h_\mathrm{top} = \lim_{n \to \infty} \frac{1}{n} \log(N(n))
\]
Substituting our spectral expansion for $N(n)$ into the definition, we get:
\[
h_\mathrm{top} = \lim_{n \to \infty} \frac{1}{n} \log \left( \sum_{i=1}^{m} S_i^2 \lambda_i^n \right)
\]
According to the Perron-Frobenius theorem, if the network is connected and non-bipartite (i.e., its adjacency matrix is non-negative, irreducible, and primitive), the largest eigenvalue $\lambda_1$ (the spectral radius) is strictly greater in absolute value than all other eigenvalues ($|\lambda_i| < \lambda_1$ for $i > 1$). Since $\lambda_1 > 0$, we can factor out its $n$-th power and rewrite the sum as: 
\[
h_\mathrm{top} = \lim_{n \to \infty} \frac{1}{n} \log \left[ \lambda_1^n \left( S_1^2 + \sum_{i=2}^{m} S_i^2 \left( \frac{\lambda_i}{\lambda_1} \right)^n \right) \right]
\]
Using the properties of logarithms, we can split this expression into two distinct terms:
\[
h_\mathrm{top} = \lim_{n \to \infty} \left[ \frac{1}{n} \log(\lambda_1^n) + \frac{1}{n} \log \left( S_1^2 + \sum_{i=2}^{m} S_i^2 \left( \frac{\lambda_i}{\lambda_1} \right)^n \right) \right]
\]
As $n$ approaches infinity, the terms $(\lambda_i / \lambda_1)^n$ exponentially decay to $0$ and the additive term $\frac{\log(S_1^2)}{n}$ also goes to zero. Therefore, the topological entropy is entirely determined by the dominant eigenvalue:
\[
h_\mathrm{top} = \log(\lambda_1)
\]

Moreover, in the context of networks, this relationship is particularly meaningful: the quantity $\lambda_1$ does not merely reflect the number of walks in the graph, but its logarithm measures the average Shannon information per step gained when observing a random walk unfold. A larger $\lambda_1$ indicates a greater number of possible distinct paths at each step, implying higher unpredictability, richer structural diversity, and overall complexity.

This connection positions $\lambda_1$ as a fundamental quantity bridging structural graph theory, information theory, and dynamics. It can be interpreted as a proxy for the network's ``intrinsic computational capacity'' or ``information-processing potential''.

Specifically, given a network with $N = 64$ nodes and an adjacency matrix whose entries are at most 2, the maximum possible $\lambda_1$ is $T = 2(N-1) = 126$, which would correspond to a fully connected weighted network; in practice, legal chess moves constrain the interactions, so this bound is never reached.

In our study, this interpretation is particularly useful: we use the spectral radius $\lambda_1$ as a global measure of the tension and complexity present on the chessboard. A higher $\lambda_1$ suggests a denser and more interconnected pattern of potential interactions between pieces (e.g., threats and defenses), and therefore a richer position in tactical possibilities and strategic depth. This justifies using $\lambda_1$ as a meaningful descriptor of the game's complexity at a given state.

\section{Validation against random and quasi-random play benchmarks}
\label{subsec:random_benchmark}
\noindent
To determine whether the characteristic ``peak and drop'' pattern observed in the strategic tension profiles of human and AI games (see Fig. \ref{fig:tension}a in the main text) is a product of structured strategic planning or merely an intrinsic, geometric property of the game of chess itself (e.g., driven solely by piece density and legal move physics), we introduce a null model benchmark. 

Furthermore, the comparison among top human players across different time constraints (Rapid/Blitz versus Classical) reveals highly similar overall profiles. While longer time limits allow slightly higher tension, both human formats exhibit a distinctly lower mid-game tension peak than elite AI engines.

\begin{figure}[!htbp]
\centering
\includegraphics[width=1\linewidth]{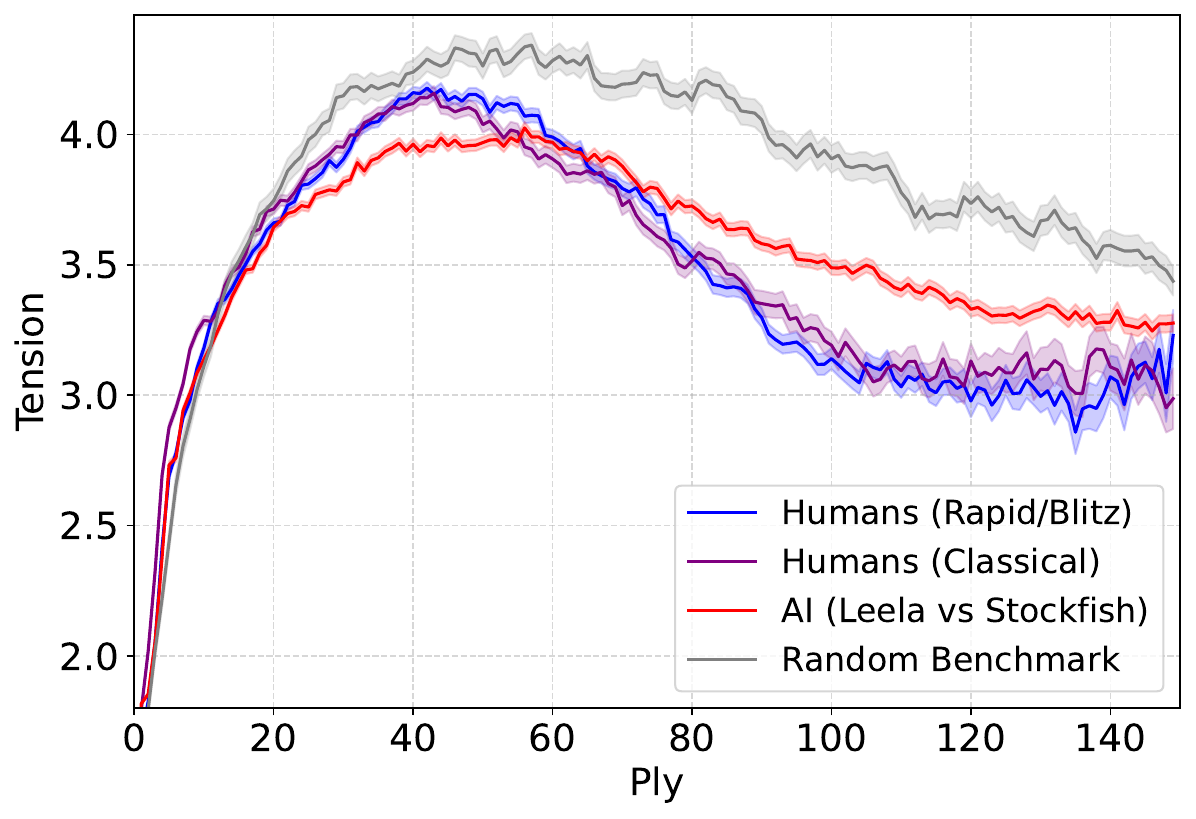}
\caption{\label{fig:random_benchmark} Comparison of the global strategic tension profiles over time (measured in plies) for human grandmasters in Rapid/Blitz formats (blue), human grandmasters under classical time controls (purple), competitive AI engines (red), and the purely random play benchmark (gray). We analyze the set of games shown in Fig.~\ref{fig:tension}. Shaded areas represent the standard error of the mean.}
\end{figure}

We simulated 300 independent chess games where both players select their moves uniformly at random from the set of all available legal moves at each ply. For each ply, we computed the global strategic tension using the same metric applied to both human and AI datasets, averaging across all simulated trajectories to obtain a baseline profile. The comparison among the tension profiles, illustrated in Fig.~\ref{fig:random_benchmark}, reveals a significantly higher tension for the random benchmark throughout the game.

While random play does exhibit an initial rise followed by a peak, its tension decays much more slowly and remains at significantly higher values compared to both human and AI games in the mid-to-late phases. Crucially, because random players lack the strategic intent to efficiently capture and remove pieces, the board remains saturated with unrefined, accidental interactions. Human and AI players, by contrast, actively resolve tactical tension through purposeful simplifications and structural liquidations. This quantitative divergence proves that the rapid drop in tension observed in real games is not an intrinsic geometric artifact of chess, but a direct signature of goal-directed strategic thinking and tactical resolution characteristic of both biological and artificial intelligence.

\adamo{The observation that the purely random baseline sustains a tension profile higher than that of elite players requires further clarification to understand under what low-level playing conditions the strategic tension escalates so drastically.}

\begin{figure}[!htbp]
\centering
\includegraphics[width=1\linewidth]{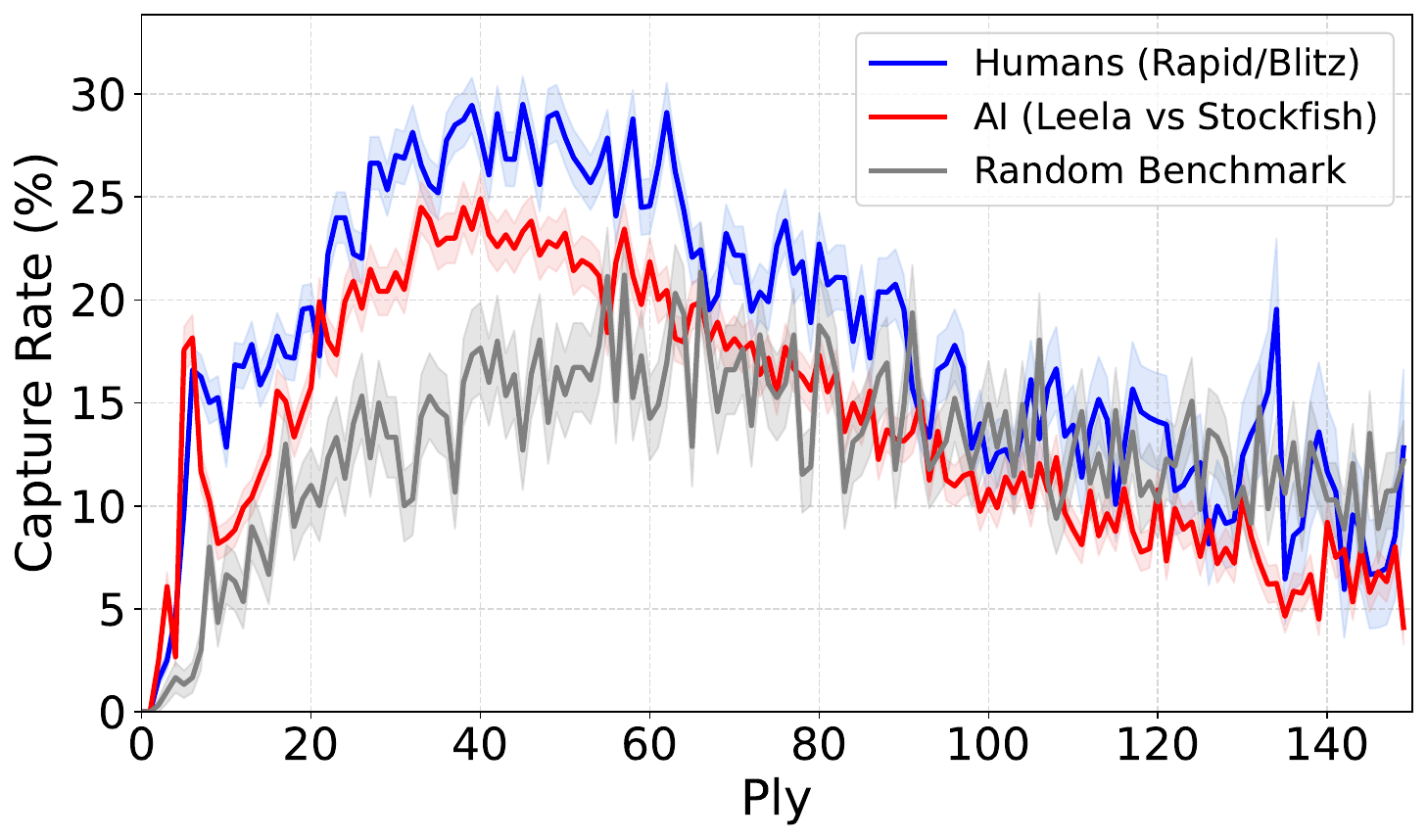}
\caption{\label{fig:capture_rates} \adamo{Empirical capture rate profiles as a function of time (measured in plies) for human grandmasters in Rapid/Blitz formats (blue), elite AI engines (red), and the purely random play baseline (gray). The curves represent the percentage of moves at each specific ply that result in a piece capture, averaged across the same game datasets analyzed in Fig.~\ref{fig:tension}. The overall global average capture rates are 20.2\% for human play, 13.0\% for AI, and 13.1\% for the purely random baseline. Shaded areas indicate the standard error of the mean.}}
\end{figure}

\adamo{Therefore, to systematically investigate how stochastic decisions shape the network topology, we introduced a hierarchy of quasi-random benchmarks. We compared the purely random baseline against ``Human-Inspired" and ``AI-Inspired" models featuring a dynamic, ply-dependent capture probability. By calibrating the capture rate at each ply to match empirical profiles from real games (illustrated in Fig.~\ref{fig:capture_rates}), this approach isolates the temporal dynamics of material simplification within an otherwise non-intentional framework. In addition to these stochastically weighted benchmarks, we analyzed the performance of the Stockfish engine under severe computational constraints, utilizing only the Hand-Crafted Evaluation (HCE) function with neural network (NNUE) evaluation disabled \footnote{Stockfish originally relied on a Hand-Crafted Evaluation (HCE) function, a traditional linear combination of human-defined heuristic features such as material balance, king safety, and piece-square tables. Modern versions achieve grandmaster-level positional understanding by replacing HCE with Efficiently Updatable Neural Networks (NNUE), which evaluate board positions via a shallow neural network trained on millions of positions. Disabling NNUE and restricting the engine to HCE allows us to isolate a more rudimentary, feature-driven form of tactical assessment.}.}

\begin{figure}[!htbp]
\centering
\includegraphics[width=1\linewidth]{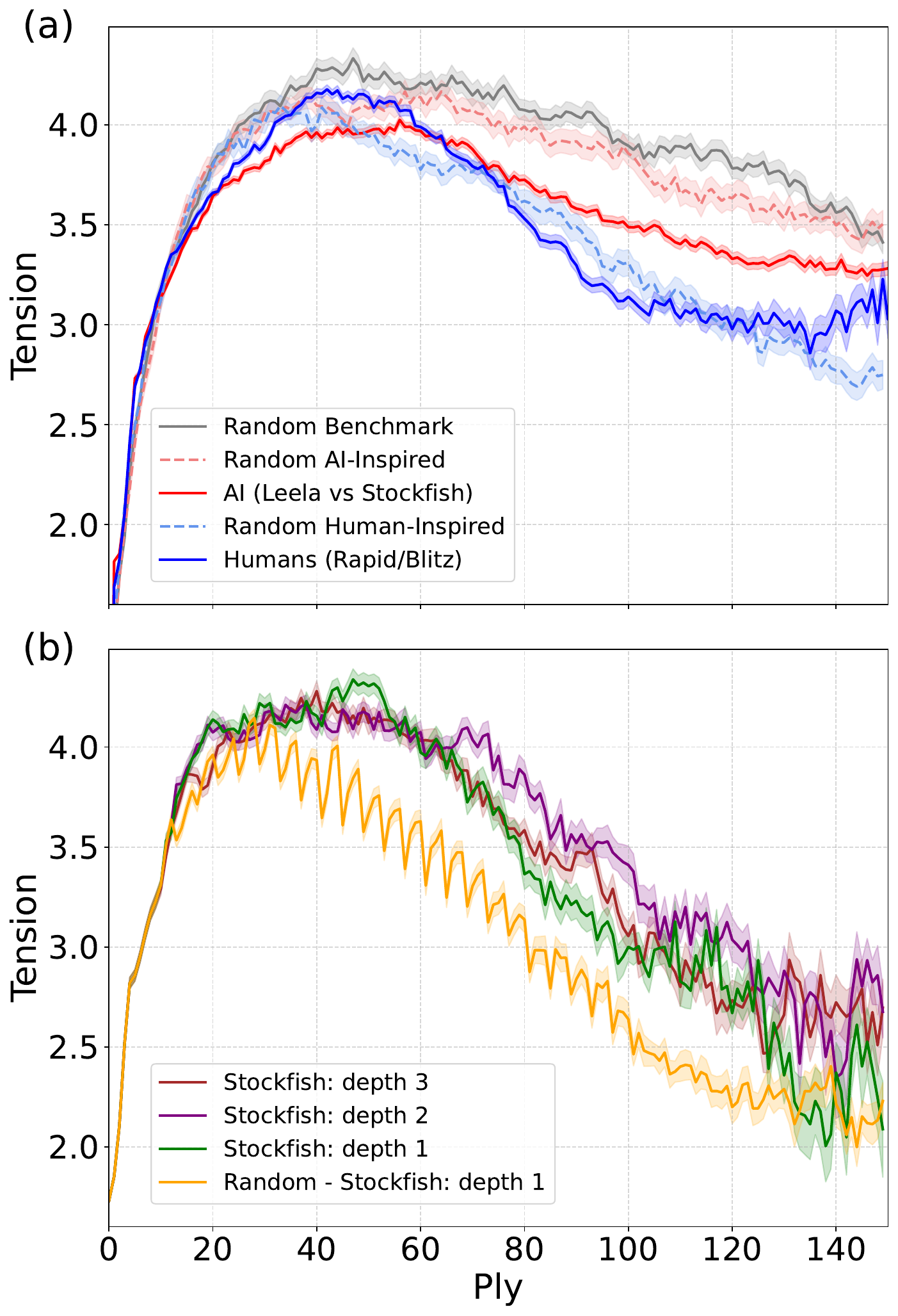}
\caption{\label{fig:extended_benchmark} \adamo{Comparison of the global strategic tension profiles over time (measured in plies) across real games, stochastic benchmarks, and constrained computational models (HCE-based Stockfish). \textbf{(a)} The top panel illustrates the tension profiles of stochastic Random Human-Inspired (dashed blue) and Random AI-Inspired (dashed red) benchmarks, which were generated using the dynamic, ply-by-ply empirical capture probabilities shown in Fig.~\ref{fig:capture_rates}. These are compared with tension profiles for human grandmasters in Rapid/Blitz formats (blue), competitive AI engines (red), and the purely random play benchmark (gray) previously shown in Fig.~\ref{fig:random_benchmark}.  \textbf{(b)} The bottom panel displays the computational frameworks, showing the deterministic Stockfish engine restricted to depth 1 (green), depth 2 (purple), and depth 3 (brown). It also features the phase-shifted alternating Random-Stockfish Depth 1 model (orange). In this mixed framework, players alternate their computational regimes at each turn: one player selects a move calculated at depth 1, while the opponent responds by selecting a move uniformly at random from the set of legal options. For each individual stochastic and computational model, a sample of 300 independent games was simulated and analyzed. Shaded areas represent the standard error of the mean.}}
\end{figure}

\adamo{The empirical results, illustrated in Fig.~\ref{fig:extended_benchmark}a, reveal a nuanced dynamical hierarchy. Within the purely stochastic models, the baseline tension is highly sensitive to the parameterization of the capture probability. Remarkably, the AI-Inspired random benchmark accumulates and sustains a tension profile only marginally lower than the pure random baseline, even though their average capture rates over the course of a game are nearly identical. Conversely, the higher capture rate of the Human-Inspired model shifts the curve downward during the mid-game, promoting a faster liquidation of pieces. Interestingly, after reaching its tension peak, the stochastic models exhibit a relatively constant negative slope, reflecting a steady, linear dissipation of tension. In stark contrast, actual games played by human experts and elite AI engines exhibit a fundamentally non-linear dissipation dynamic. Following their respective tension peaks, real games undergo a rapid, sharp collapse in tension. This steep decline subsequently transitions into a much slower decay, ultimately settling into a near-constant tension plateau after ply 120.}

\adamo{A fundamental regime shift occurs as soon as even a minimal degree of goal-directed optimization is introduced (see Fig.~\ref{fig:extended_benchmark}b). The deterministic Stockfish Depth 1 engine demonstrates that a deep, complex evaluation is not strictly required to collapse the network complexity. This shallow optimization is already sufficient to significantly depress the overall tension profile, limiting early accumulation and forcing a distinct post-peak decay—a pattern that remains largely consistent for Depth 2 and Depth 3, which exhibit only marginally higher tension values.}

\adamo{Qualitatively, while Depth 1 acts as a purely greedy, hyper-reactive agent with no awareness of opponent retaliation, Depths 2 and 3 introduce basic threat-evasion and tactical validation. However, all three models share a fundamentally myopic computational horizon: they act as strict local optimizers that instinctively resolve imbalances and liquidate material rather than sustaining long-term strategic tension.} 

\adamo{Most intriguingly, the mixed ``Random - Stockfish: Depth 1" model (orange curve) exhibits a structurally lower tension profile throughout the mid-game compared to pure Depth 1. This emergent behavior indicates that injecting stochasticity into the game loop allows the greedy, short-sighted evaluation of a Depth 1 agent to ruthlessly exploit tactical blunders, accelerating material liquidation and rapidly draining the interaction network. Furthermore, the distinct oscillatory behavior observed in the tension profile directly reflects the turn-by-turn alternation between goalless random moves and goal-directed Depth 1 responses. Ultimately, these findings demonstrate that while game geometry dictates the potential for tension, the mere presence of minimal, localized strategic intent is sufficient to trigger tactical resolution and to dictate the velocity of its dissipation.}

\adamo{To formalize this and determine the impact of strategic intent, we introduce a continuous probabilistic model controlled by parameter $p \in [0, 1]$, representing the probability of executing a goal-directed move (Depth 1) versus a random one ($1-p$).}

\begin{figure}[!htbp]
\centering
\includegraphics[width=1\linewidth]{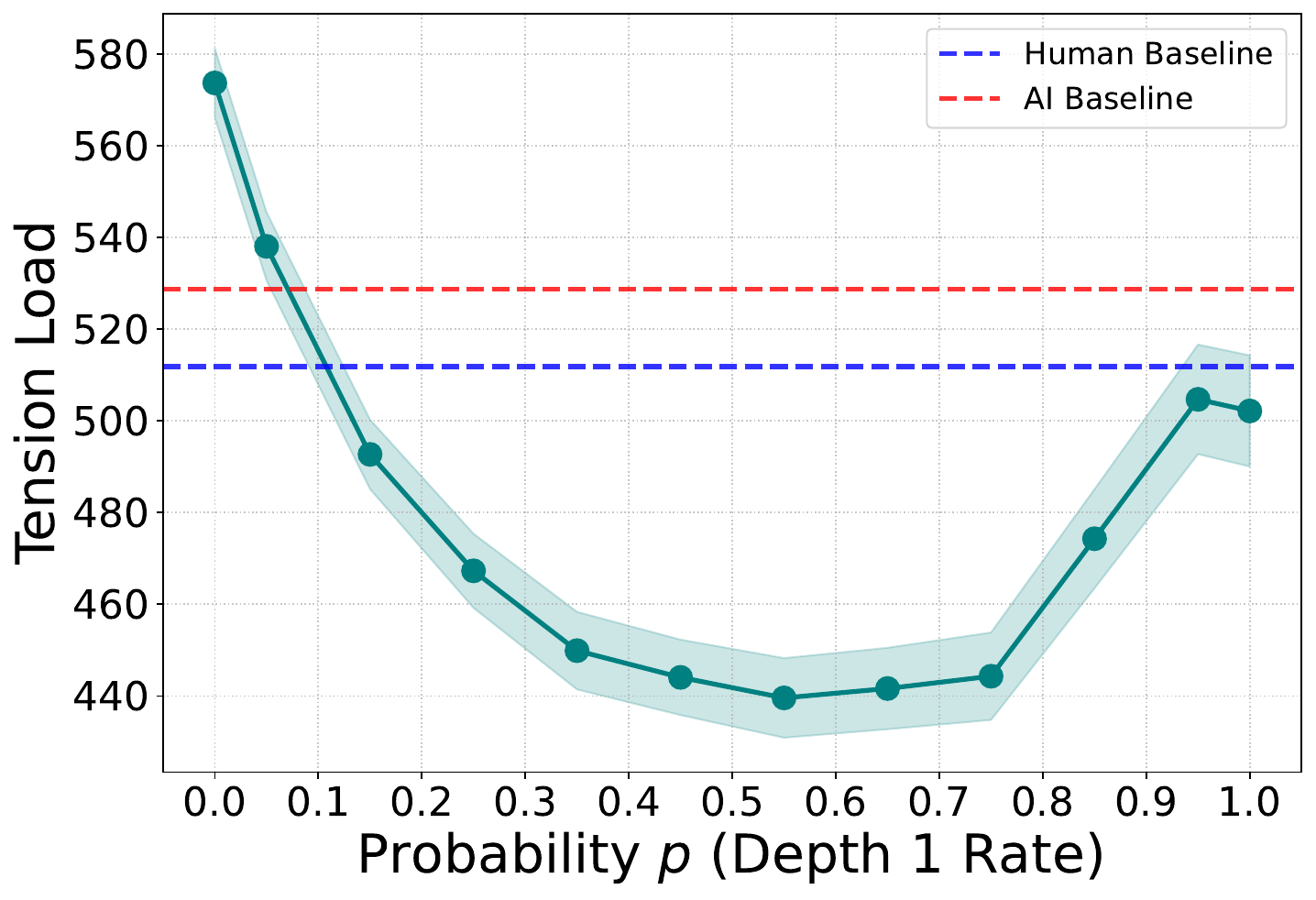}
\caption{\label{fig:phase_transition} \adamo{Tension load as a function of the optimization probability $p$, defined as the probability of executing a goal-directed move (HCE-based Stockfish at Depth 1) rather than a purely random one ($1-p$). The baseline ($p=0.0$) represents pure random play, while $p=1.0$ represents pure Depth 1 evaluation. Shaded areas indicate the cumulative standard error of the mean.}}
\end{figure}

\adamo{The results of tension load (sum of the tension profile over $N = 150$ plies), depicted in Fig.~\ref{fig:phase_transition}, reveal a non-monotonic behavior. The pure random baseline ($p=0.0$) exhibits the highest tension load, characterizing an entropic regime of unchecked interactions. However, injecting even a small fraction of tactical optimization results in a sharp collapse in structural tension.}

\adamo{Crucially, the system does not simply decay toward the deterministic baseline ($p=1.0$). Instead, the curve exhibits a distinct global minimum between $p \approx 0.4$ and $p \approx 0.7$, before rising again. This demonstrates that a mixed stochastic-evaluative regime is the most destructive to network complexity. The random moves constantly inject catastrophic tactical blunders into the system, which the hyper-reactive Depth 1 agent instantly exploits. This mechanism acts as an accelerated liquidation engine, draining the interaction network much faster than a pure Depth 1 agent ($p=1.0$), which, without a continuous supply of random blunders to exploit, sustains a slightly higher baseline of tension.}

\section{Example of tension profile for a single game}
\label{subsec:example_tension}
\begin{figure}[!htbp]
\centering
\includegraphics[width=0.9\linewidth]{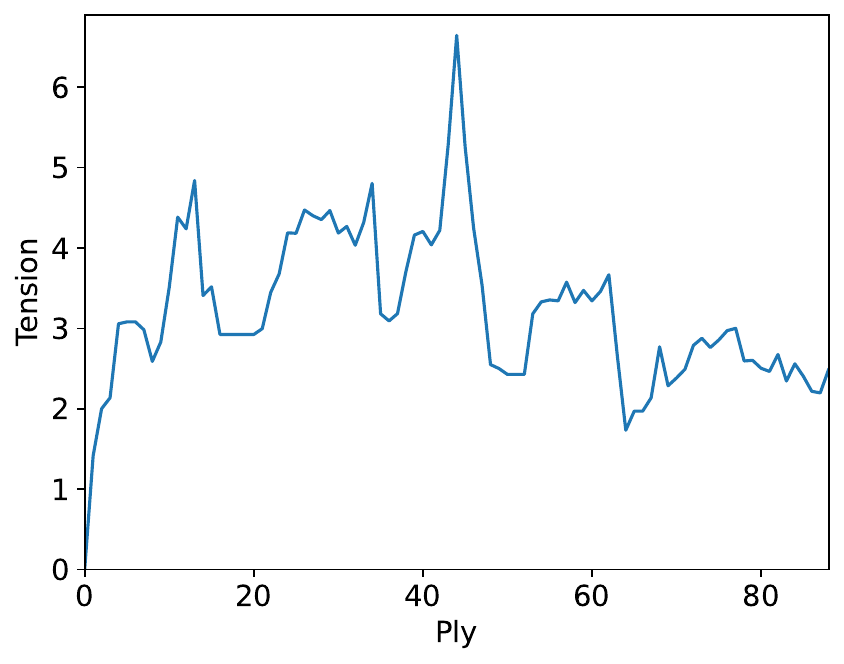}
\caption{\label{fig:single_game} Tension profile of a single game between two top grandmasters: Magnus Carlsen (White) vs. Alexey Sarana (Black), in the Round 6 of the Titled Tuesday tournament on chess.com, 26 November 2024.}
\end{figure}

\noindent
In this section, we present the tension profile of a single game played between top grandmasters (Fig.~\ref{fig:single_game}). Without averaging across multiple games, pronounced fluctuations become apparent: individual moves can significantly disrupt the network of interactions among the pieces, leading to sudden changes in tension. This highlights how highly sensitive the system is to local positional variations, with abrupt shifts reflecting critical strategic or tactical decisions.

The moves of the game are listed below: 1. c4 e6 2. Nc3 d5 3. d4 c5 4. cxd5 exd5 5. Nf3 Nc6 6. Bg5 Be7 7. Bxe7 Ngxe7 8. e3 c4 9. Be2 O-O 10. O-O a6 11. a4 Bf5 12. b3 Qa5 13. Rc1 b5 14. axb5 axb5 15. bxc4 bxc4
16. Nd2 Nb4 17. e4 Bxe4 18. Ncxe4 dxe4 19. Nxe4 Nbc6 20. Bxc4 Rad8 21. d5 Nxd5 22. Ra1 Nde7 23. Rxa5 Rxd1 24. Rxd1 Nxa5 25. Ba2 Nac6 26. g3 g6 27. Rd7 Ra8 28. Bb3 Ra3 29. Nc5 Ra5 30. Ne4 Ra3 31. Rb7 Ra7 32. Rxa7 Nxa7 33. Ng5 Nac6 34. Bxf7+ Kg7 35. Ba2 h6 36. Ne4 Nd4 37. f4 Nef5 38. Kf2 Ne7 39. g4 Nb5 40. Ke3 Nc6 41. f5 gxf5 42. gxf5 Nc7 43. f6+ Kg6 44. Kf4 Nd8 45. Ke5.

The peak tension is observed at ply 44, with its interaction network illustrated in Fig.~\ref{fig:network}.

\section{Maximized tension game}
\label{subsec:max_tension}
\begin{figure}[!htbp]
\centering
\includegraphics[width=1\linewidth]{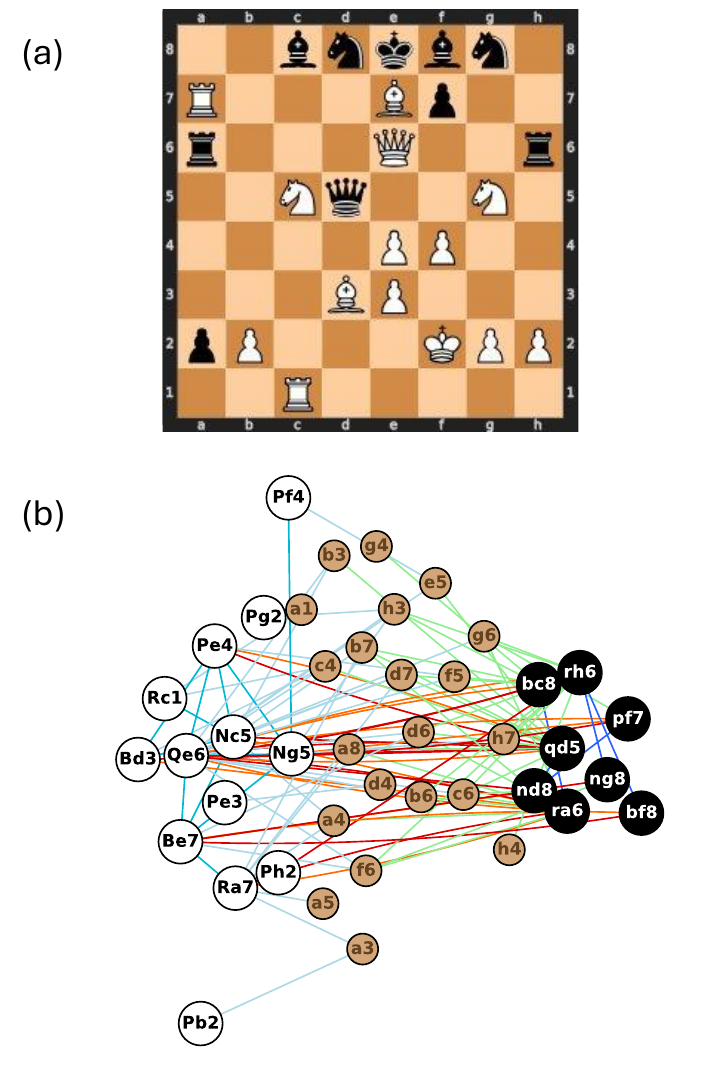}
\caption{\label{fig:network_max} Tension network obtained by maximizing tension move after move. (a) Chessboard at ply 44, generated by choosing the legal move that maximizes the tension at each ply. While this procedure yields a position that is strategically nonsensical from a conventional chess perspective, it illustrates the possible theoretical upper bound of the interaction network. (b) The corresponding network represents the interactions on the chessboard, using the same color and edge scheme as in Fig.~\ref{fig:network}.}
\end{figure}

\noindent
If we consider the unrealistic and absurd game in which, at each ply, the players randomly choose one of the moves that maximizes tension, by ply 52 the tension reaches $T \simeq 10.4$ (Fig.~\ref{fig:network_max}).

The first 60 moves of the game are listed below: 1. d3 h6 2. Bg5 e5 3. Be7 Nc6 4. Qd2 d5 5. Qxh6 Qd6 6. Qe6 Nd8
7. c4 Rh6 8. cxd5 e4 9. Nc3 g5 10. Na4 b5 11. Rc1 Rb8 12. Rxc7 Rb7 13. Nc5 Qxd5 14. dxe4 Rb6 15. e3 Ra6 16. Rxa7 b4 17. Bd3 Bg7 18. Nf3 Bf8
19. Nxg5 Bd7 20. f4 Bc8 21. Kf2 b3 22. Rc1 bxa2 23. Rc3 a1=Q 24. Ra3 Qd1 25. Ra1 Qg4 26. Ra4 Qgf5 27. Kg3 Qfxe4 28. Ra8 Qef5 29. Ra7 Bb7 30. Kf2 Bc8.

After ply 52, the game maintains a relatively constant level of tension indefinitely, indicating that the game ceases to progress meaningfully.

\section{Structural robustness of the tension metric}
\label{subsec:Ablation_Study_Tension}
\noindent
To verify that the observed tension profiles are not an artifact of combining distinct interaction types (attacks, defenses, and empty square controls), we performed a structural ablation study. We compared the full interaction network against two restricted sub-networks: one isolating attack and defense links, and a minimal one consisting solely of attack links.

\begin{figure}[!htbp]
\centering
\includegraphics[width=1\linewidth]{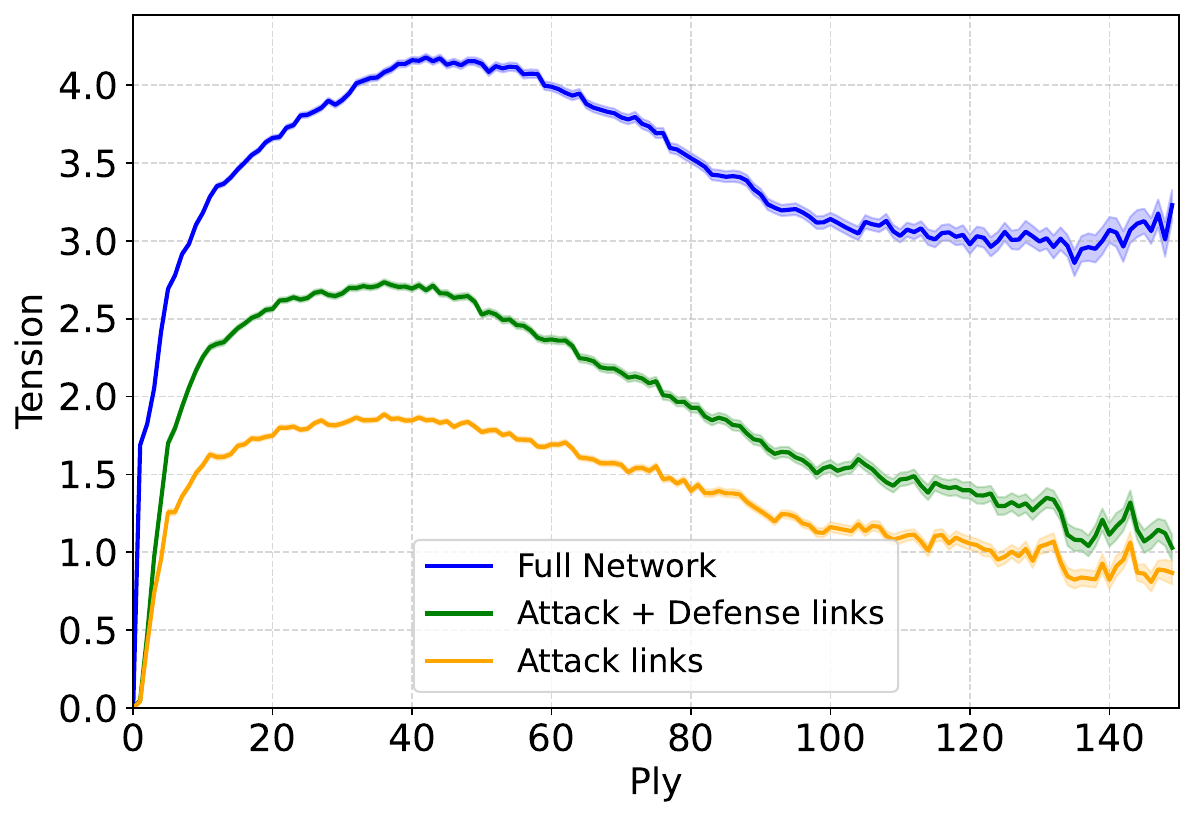}
\caption{\label{fig:ablation_tension} Structural robustness of the tension metric across different network topologies. The figure illustrates the average strategic tension as a function of the ply, computed for human games (those analyzed in Fig.~\ref{fig:tension}). Three network configurations are compared: the full interaction network (blue line), which includes attacks, defenses, and empty-square controls; a deconstructed sub-network restricted to attack and defense links (green line); and a minimal sub-network consisting solely of attack links (orange line). Shaded regions denote the standard error of the mean.}
\end{figure}

As shown in Fig.~\ref{fig:ablation_tension}, while sparser networks naturally yield lower tension values, the macroscopic trajectories remain qualitatively identical. Across all configurations, tension exhibits a steady ascent towards a midgame peak followed by a sustained decline. This structural robustness confirms that the fundamental pattern of tactical escalation is preserved across different topologies.

\section{Tension peak analysis}
\label{subsec:tension_peak}
\begin{figure}[!htbp]
\centering
\includegraphics[width=1\linewidth]{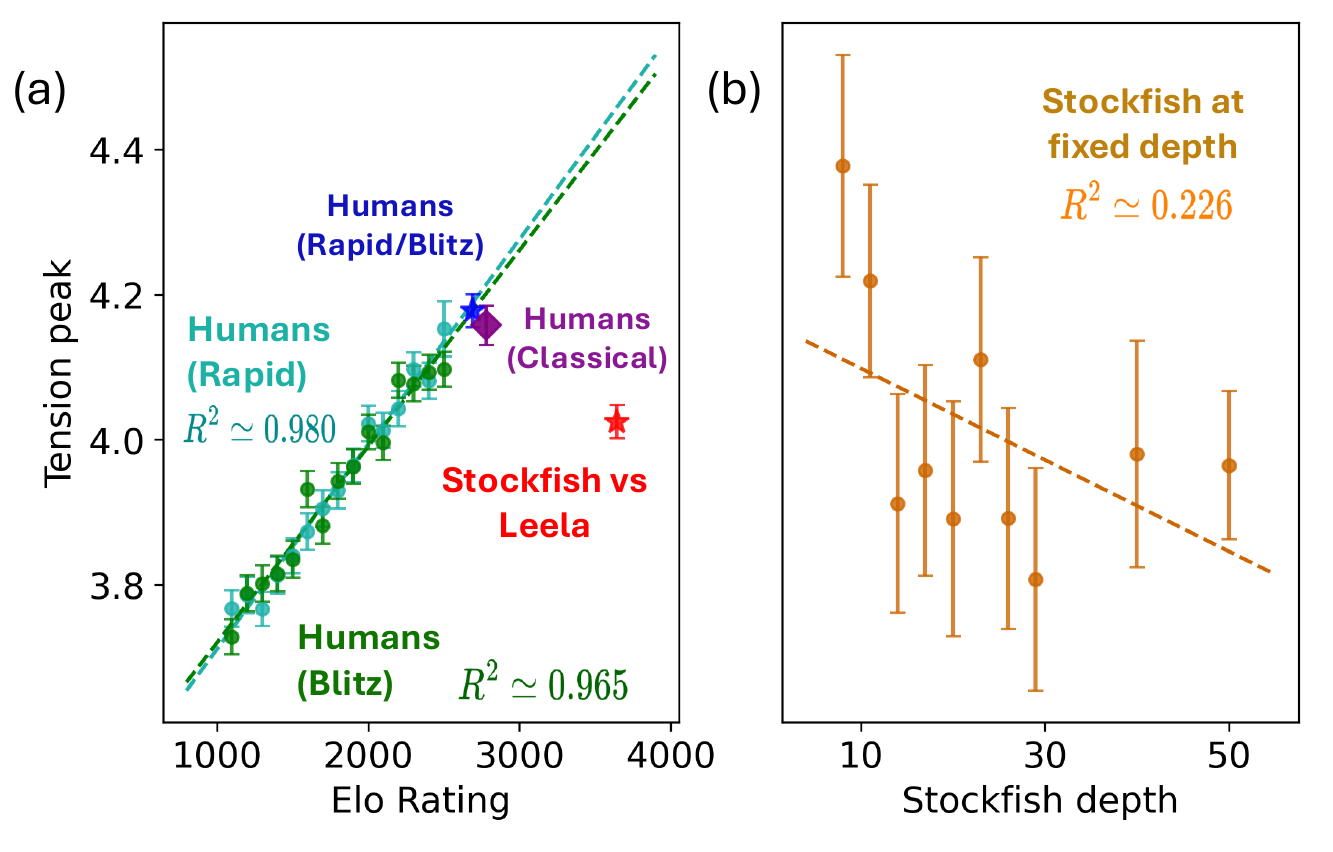}
\caption{\label{fig:Tension_peak} Tension peaks for different Elo ratings and Stockfish depths. (a) Tension peaks are shown for rapid games (turquoise dots) and blitz games (green dots). The plot also includes the peaks of top human players in rapid/blitz (blue star) and classical games (purple diamond), as well as elite AI matches between Stockfish and Leela from official TCEC tournaments (red star). All data points are plotted against their estimated Elo ratings. Each dot/star represents the average across 1,200 games. The blue and red stars were derived from the data analyzed in Fig.~\ref{fig:tension}. For reference, blitz games are played with very short time controls (typically 3–5 minutes per player), rapid games with intermediate time controls (around 10–25 minutes per player), and classical games with long time controls (often 60 minutes or more per player), allowing deeper strategic play. (b) We represent tension peaks from games between versions of Stockfish at fixed depth levels (orange dots), computed from a smaller sample of 120 games per dot. The two panels also show the $R^2$ values of the corresponding linear regressions.}
\end{figure}

\noindent
To investigate the origins of the pronounced tension peak observed in the early stages of human games, we designed an empirical test (see Fig.~\ref{fig:Tension_peak}) to compare how the maximum tension peak scales with human skill (Elo rating) versus pure computational power (engine depth). 

As illustrated in Fig.~\ref{fig:Tension_peak}a, the maximum tension peak in human games exhibits a clear linear growth with respect to the players' Elo rating. This upward trend is consistent across both rapid and blitz time controls, ultimately culminating in the extreme peaks recorded by elite grandmasters in classical and rapid/blitz formats. 

In stark contrast, evaluating engine play reveals an opposing dynamic. Fig.~\ref{fig:Tension_peak}b demonstrates that increasing the computational power (Stockfish depth) does not generate higher tension peaks; rather, it leads to a slight decrease in the maximum tension.

This divergence strongly supports the hypothesis that the extreme early-game tension observed in top-level human play is not a mathematical necessity for optimal chess. Instead, it is primarily a byproduct of modern opening preparation. As human skill increases, players memorize deeper, highly critical theoretical lines, deliberately navigating into and sustaining immense tactical complexity to maximize the opponent's cognitive load. Conversely, engines evaluate and resolve these positions through objective calculation, naturally diffusing early tension without the psychological need to test an opponent's theoretical preparation.

\section{Additional properties of the chessboard and network}
\label{subsec:additional_properties}
\begin{figure}[!htbp]
\centering
\includegraphics[width=1\linewidth]{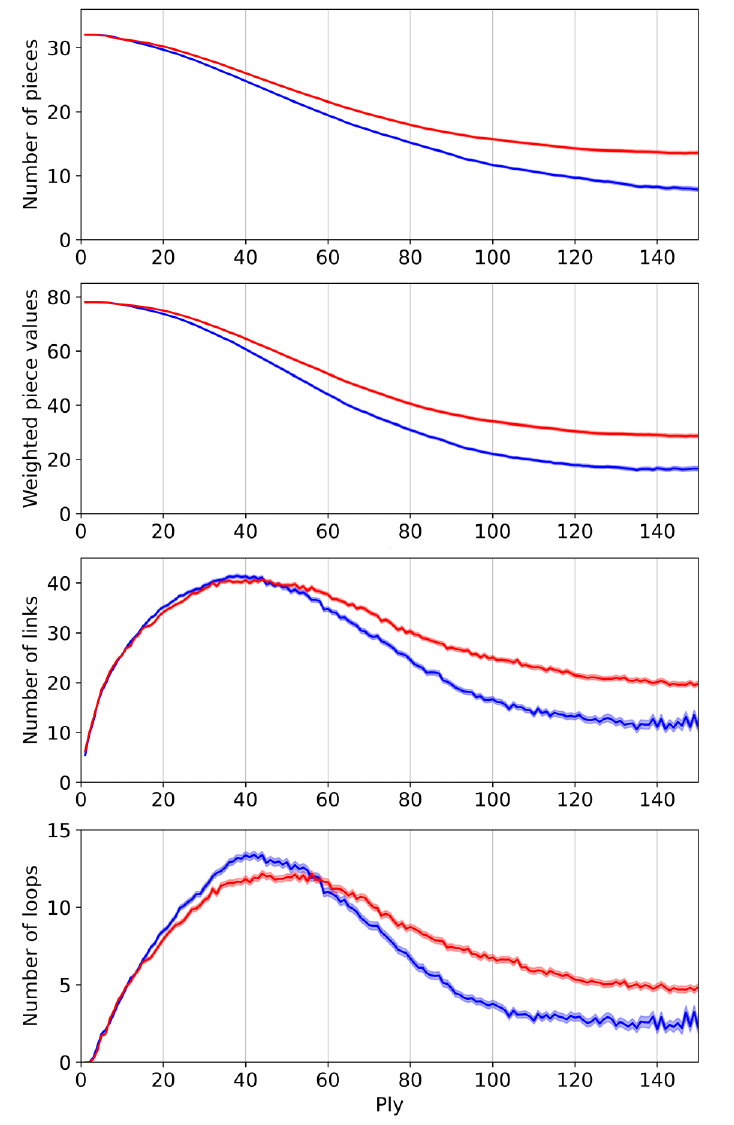}
\caption{\label{fig:properties_supp} Comparison of additional structural properties across human and AI games. We analyze the set of games shown in Fig.~\ref{fig:tension}. Four measures are considered: the number of pieces, the weighted material, the number of links, and the number of loops in the interaction network. The color scheme is red for AI games and blue for human games, respectively.}
\end{figure}

\noindent
In this section, we present further comparative metrics between human and AI chess games, including the number of pieces, the sum of the weighted material in the chess board - where pieces are valued at 1 (pawns), 3 (knights and bishops), 5 (rooks), 9 (queens), and 0 (kings) - along with the number of links and the number of loops in the tension network. These measures provide complementary insights into positional advantages and dynamic board conflicts across different types of games (see Fig.~\ref{fig:properties_supp}). 

\begin{figure}[!htbp]
\centering
\includegraphics[width=1\linewidth]{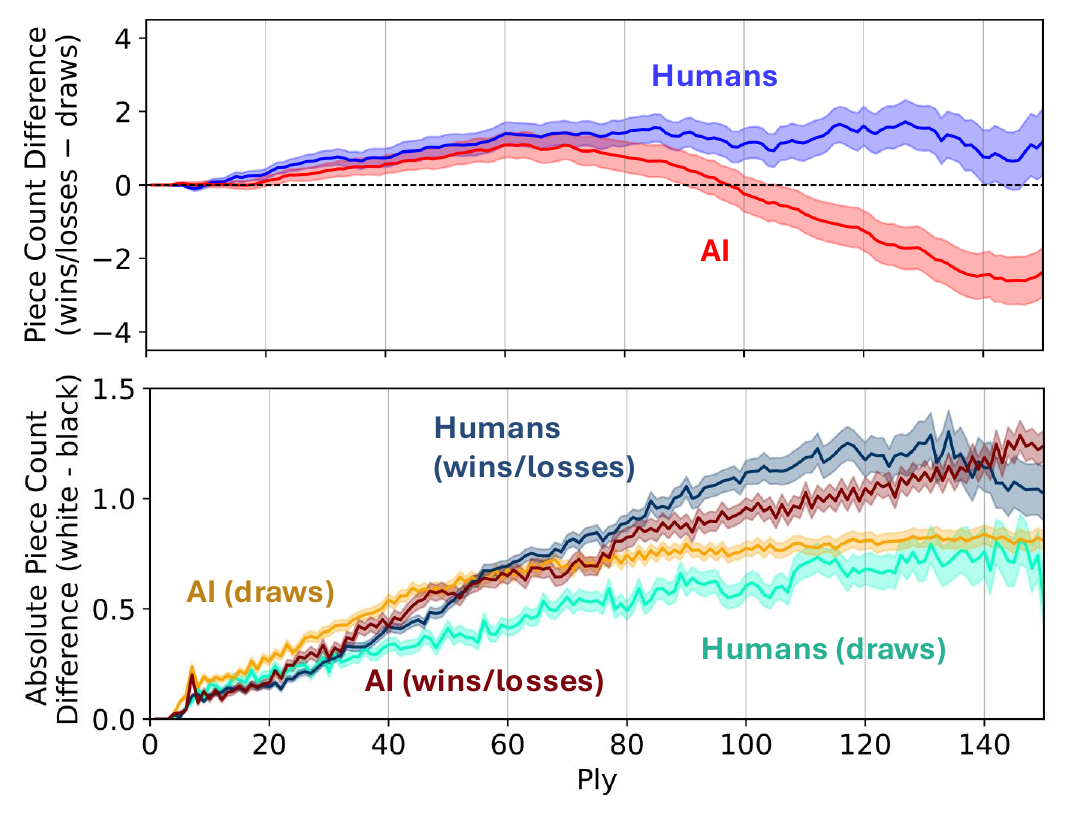}
\caption{\label{fig:Piece_Count_Difference} Analysis of piece count differences. We consider the same set of games shown in Fig.~\ref{fig:tension}. In the first panel, for both human and AI games, we report the difference in the number of pieces between games that end in wins or losses and those that end in draws. Colors distinguish AI games (red) from human games (blue). In the second panel, we report the absolute difference in the number of pieces between the two competing players. This analysis is repeated for all categories shown in Fig.~\ref{fig:tension}, using the same color scheme.}
\end{figure}

Moreover, we also analyzed the difference in the number of pieces between games ending in wins or losses and those ending in draws, separately for human and AI matches, to assess how material imbalances are associated with decisive outcomes across different cases (Fig.~\ref{fig:Piece_Count_Difference}). In addition, we examined the absolute difference in the number of pieces between the two competing players, providing a direct measure of material asymmetry within each game.

\section{Stockfish evaluation for humans and engines}
\label{subsec:stockfish_evaluation}
\begin{figure}[t]
\centering
\includegraphics[width=1\linewidth]{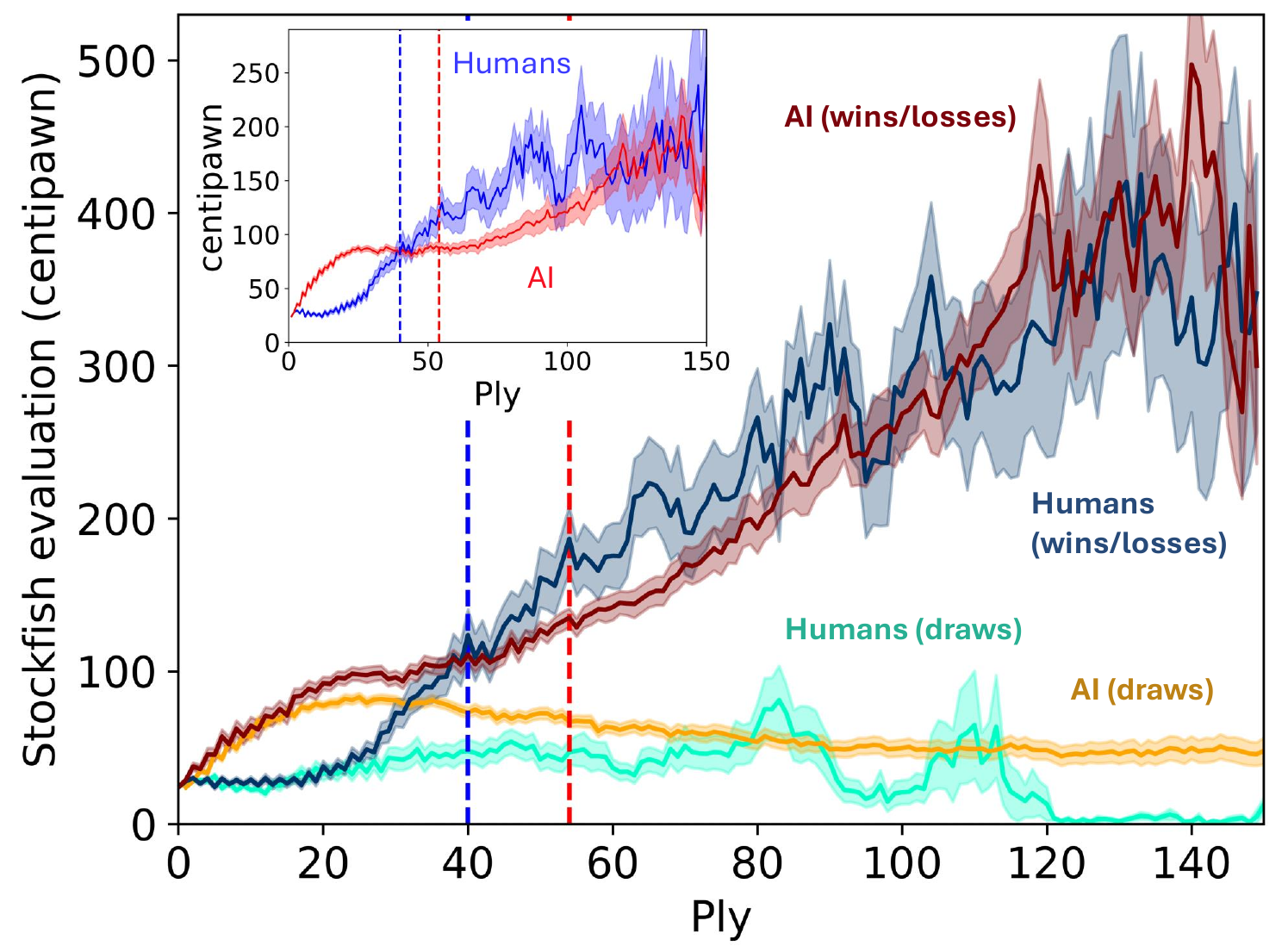}
\caption{\label{fig:stockfish_eval} Stockfish evaluation during games for humans and AI. We consider 120 games for each (a subset of those analyzed in Fig.~\ref{fig:tension}), with evaluations computed at a fixed depth of 20. The same color scheme as in Fig.~\ref{fig:tension} is used: light blue and blue denote humans, whereas orange and red denote AI, with light and dark shades distinguishing draws from decisive games. We also show the overall average evaluations, along with vertical dashed lines indicating the ply corresponding to the maximum tension for both groups, humans and AI, shown in blue and red, respectively.}
\end{figure}

\noindent
To better analyze the progression of games, we also examined the Stockfish evaluation, again distinguishing between games played by humans or AIs, and whether they ended in a draw or not (Fig.~\ref{fig:stockfish_eval}). The evaluation is a numerical score, measured in centipawns (hundredths of a pawn), that represents the advantage in a chess position, with positive for White, negative for Black, and near zero for equality. Specifically, in figure, we report the absolute value of the Stockfish evaluation, thereby capturing the magnitude of the advantage irrespective of which side (White or Black) is winning.

What emerges is that games between AIs show a rapid increase in the absolute value of the Stockfish evaluation from the very first plies. As the game progresses, the average evaluation values begin to diverge depending on the eventual outcome: games that end in a win or loss exhibit a steady increase in evaluation with each ply, whereas those that result in a draw maintain relatively low evaluation values throughout.

In contrast, in human games, the evaluation remains near zero during the opening phase. It is worth noting that, in fact, it was only starting from 2023 that official rules began allowing Leela Chess Zero to select its openings directly from a database. However, as the game progresses, a similar divergence in evaluation can be observed. Notably, this shift typically occurs around ply 40, when tension generally peaks, marking a transition in which evaluations for decisive games begin to rise consistently. In contrast, those for drawn games remain comparatively stable. Barthelemy~[\onlinecite{barthelemy2025fragility}] has already observed this result, using a similar framework for capturing the notion of fragility within chess games.

In human games, the average tension values exhibit greater fluctuations, likely due to the lower predictability of human play and, more specifically, the greater variability across individual games, especially when different players are involved. Moreover, when focusing on tension values in drawn games, a notable difference emerges: in AI matches, tension decreases towards zero much more gradually. This is mainly because engine games are, on average, longer than human games.

To better understand the type of information conveyed by the Stockfish evaluation, we analyzed 2,400 games (the same subset used to reproduce Fig. \ref{fig:tension}) to compute the probability that a player will go on to win the game when holding an evaluation advantage of at least 100 centipawns at ply 40. In human games, this probability is approximately 71 \%, whereas in AI games, it is approximately 59 \%.

To rigorously investigate the temporal relationship between the emergence of a decisive advantage and the close profile in strategic tension, we performed a time-aligned event study. Averaging tension across absolute ply counts can sometimes blur specific tactical turning points due to varying lengths and different opening phases across individual games. To overcome this, we synchronized the tension trajectories of decisive games based on a specific, objective critical event.

For each decisive game, we evaluated the position at each ply using Stockfish at a fixed depth of 20. We identified $t^*$ as the exact ply at which the absolute algorithmic evaluation first crosses the 100-centipawn threshold. This threshold was chosen because it typically represents the critical moment when a clear, structural advantage is established on the board. We then extracted a symmetric time window of 40 plies around this divergence point (from $t^* - 20$ to $t^* + 20$) and computed the mean tension and its standard error across all valid games at each relative ply.

\begin{figure}[!htbp]
\centering
\includegraphics[width=1\linewidth]{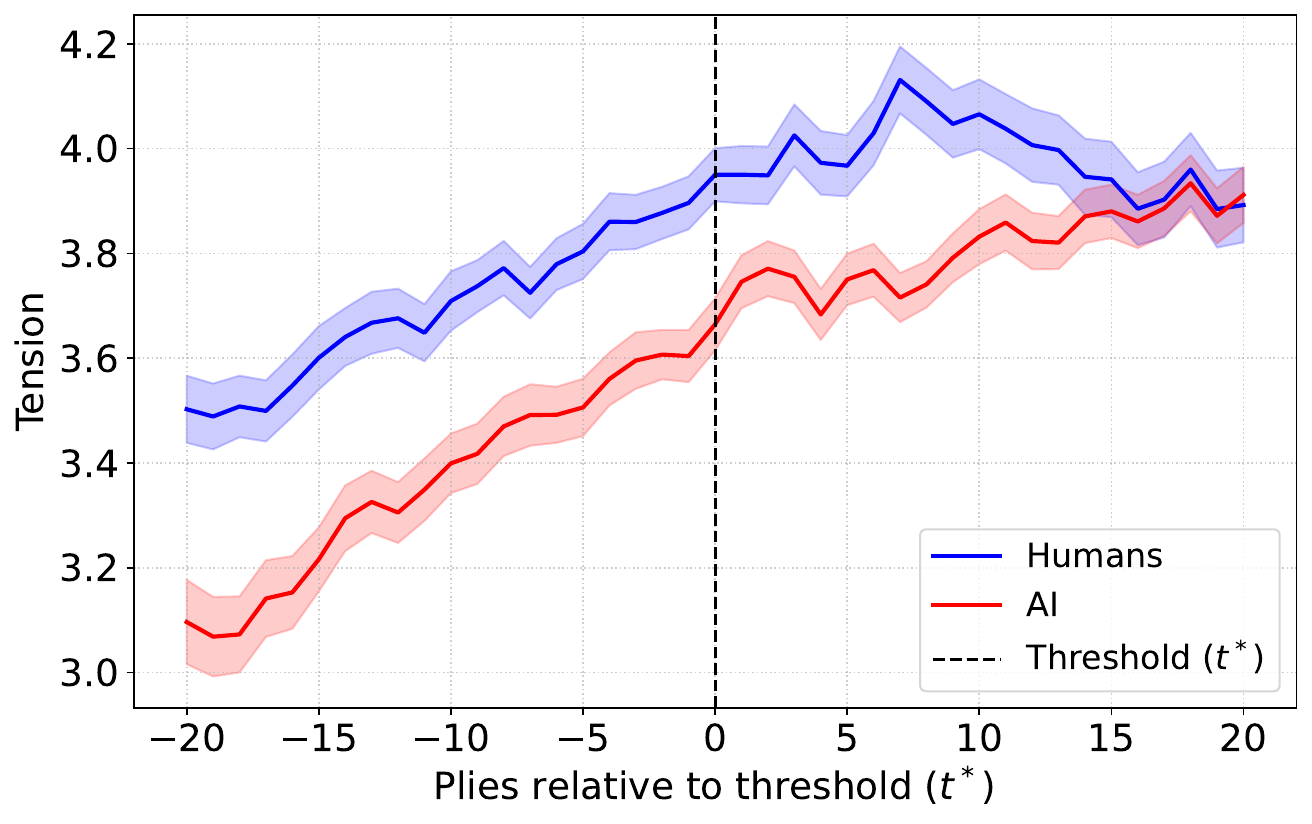}
\caption{\label{fig:Event_Study} Time-aligned event study of strategic tension for decisive human and AI games. We consider 200 games for each (a subset of those analyzed in Fig.~\ref{fig:tension}). The trajectories are synchronized at $t^* = 0$ (vertical dashed line), defined as the first ply where the absolute Stockfish evaluation exceeds 100 centipawns. The solid lines represent the average tension for Humans (blue) and AI (red), with shaded areas indicating the standard error of the mean.}
\end{figure}

As shown in Fig.~\ref{fig:Event_Study}, aligning the games at $t^*$ reveals a highly consistent structural pattern for both human and engine play. In the plies leading up to the threshold ($t < 0$), strategic tension steadily increases as the players navigate complex, unresolved positions. Shortly after passing $t^*$, the tension gradually reaches its peak before initiating a sustained decline. This pattern is particularly pronounced in human games, where the peak occurs noticeably earlier than in AI games.

This temporal ordering highlights that a player's decisive advantage is unlikely to stem from outmaneuvering the opponent during the absolute peak of complexity; rather, this pivotal advantage is most often established beforehand, during the tension's ascent. Once a clear advantage emerges (at $t^*$) and the immediate tactical crisis begins to resolve, the game generally transitions into a conversion phase. This phase is naturally characterized by piece trades and a steady reduction of topological complexity as the winning side seeks to secure the result and limit counterplay.

\section{Correlation between tension and the number of chess pieces}
\label{subsec:tension_number_pieces}
\begin{figure}[!htbp]
\centering
\includegraphics[width=0.9\linewidth]{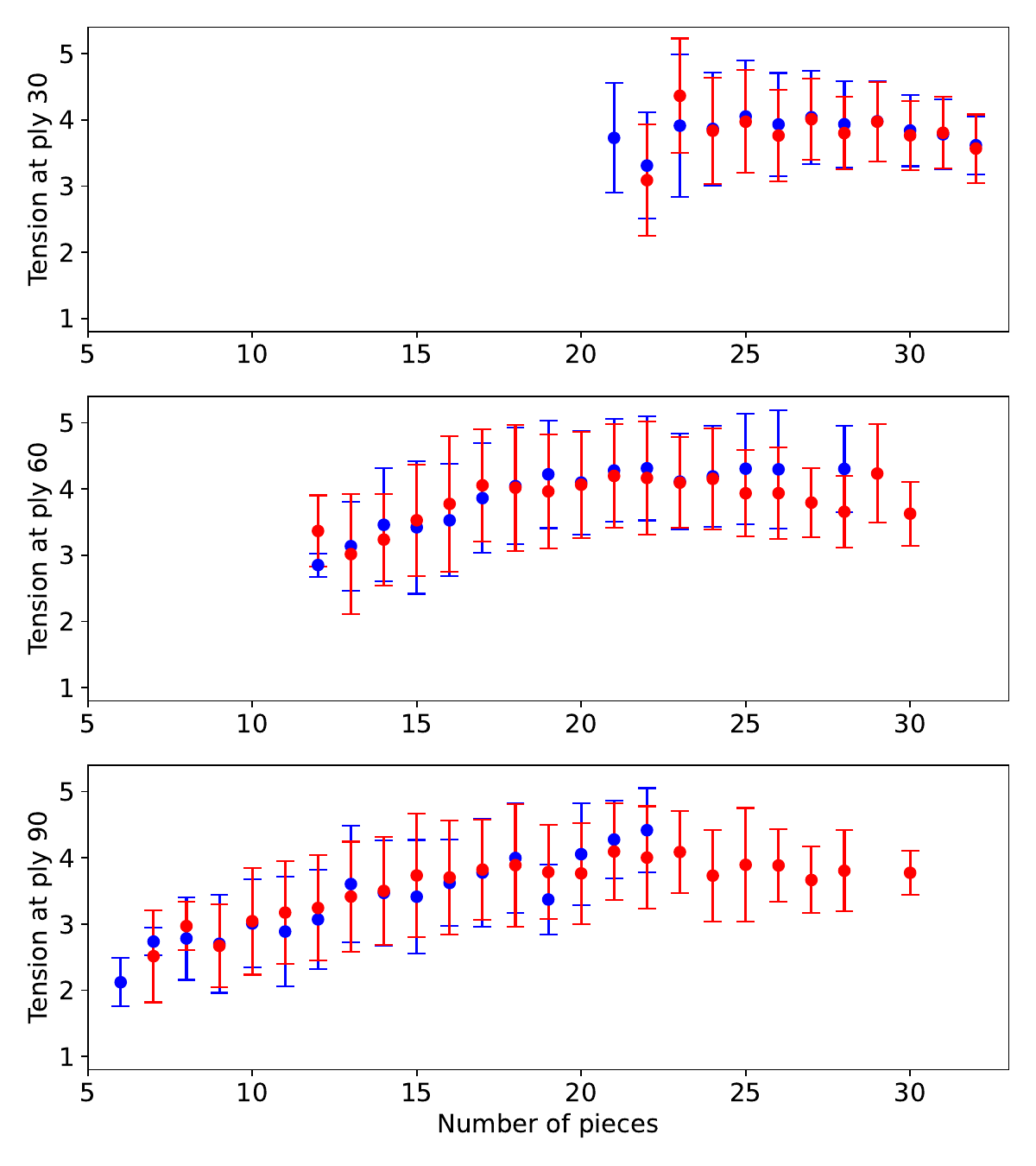}
\caption{\label{fig:tension_pieces} Analysis of the dependence of tension on the number of chess pieces at three game moments. We analyze the set of games shown in Fig.~\ref{fig:tension}. Average tension values are shown for each number of chess pieces, distinguishing between three game moments: ply 30, 60, and 90. The color scheme is red for AI games and blue for human games, respectively. The error bars correspond to the standard deviation calculated for each represented value.}
\end{figure}

\noindent
We investigated the correlation between the measured tension values and the number of chess pieces present on the board. Naturally, this correlation depends heavily on the stage of the game being considered. For this reason, we analyzed three distinct moments in the game, specifically at ply 30, 60, and 90, to show how, on average, tension varies with the number of pieces on the board, both for the AI and for human players (Fig.~\ref{fig:tension_pieces}).

We observe that, on average, tension increases with the number of chess pieces across all three stages of the game. However, this increase diminishes rapidly as the number of pieces increases, eventually approaching a roughly constant value—or even decreasing slightly. This may be because when the number of pieces is significantly higher than what would typically be expected at a given ply, it might indicate that no major confrontation has yet occurred. Such confrontations often lead to increased tension as more pieces are brought into key positions on the board. Moreover, an excessive number of pieces on the board can severely limit their mobility, reducing their ability to control important areas or pose threats to the opponent. Therefore, it is reasonable to expect that, to maximize tension, there exists an ideal range for the number of pieces — neither too high nor too low.

\section{Skill-calibrated comparison of human and synthetic players}
\label{subsec:Skill-Calibrated_Comparison}
\begin{figure}[!htbp]
\centering
\includegraphics[width=1\linewidth]{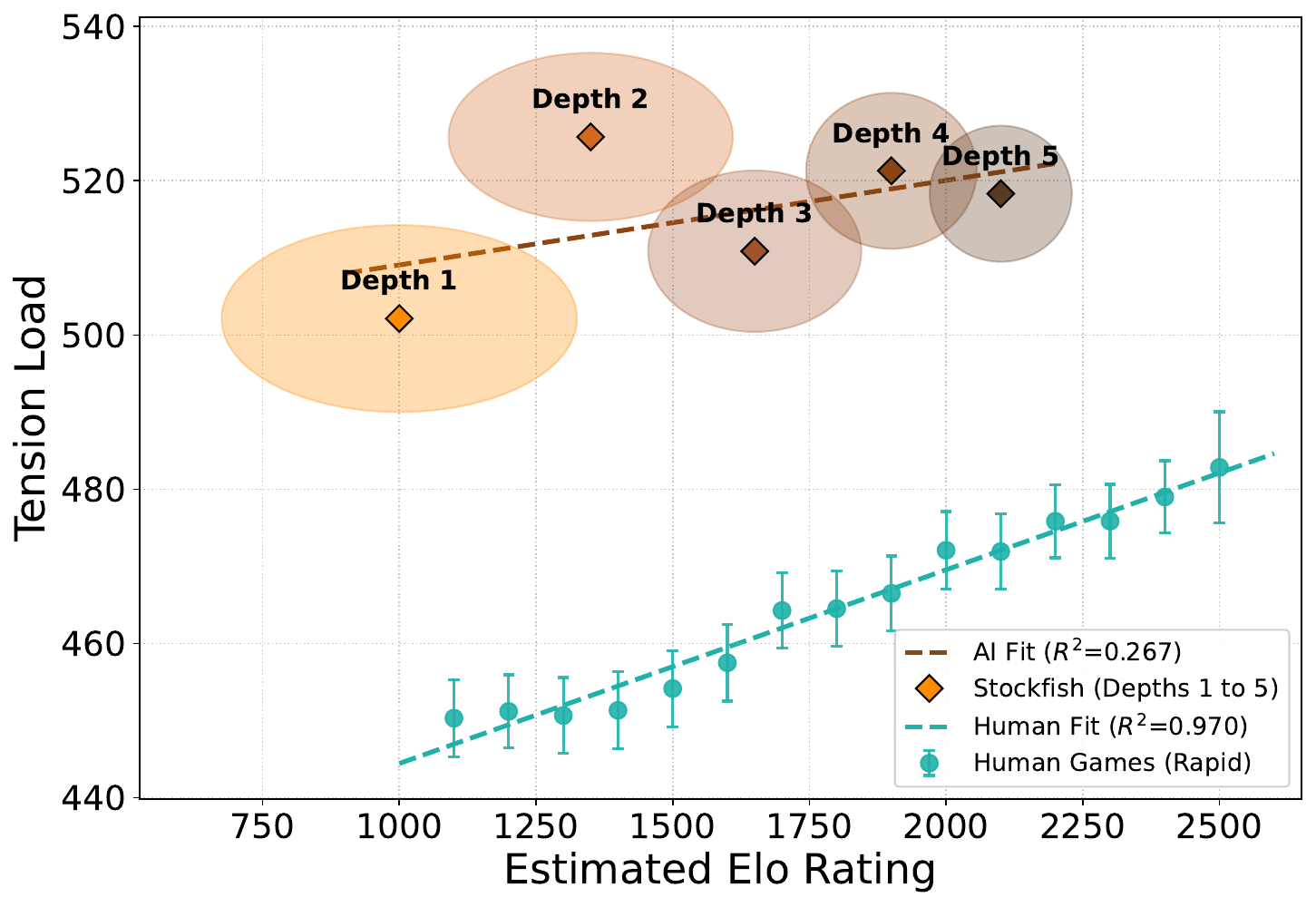}
\caption{\label{fig:Skill-Calibrated} \adamo{Comparison of the tension load between human players (Rapid games analyzed in Fig.~\ref{fig:tension_load}) and synthetic agents (Stockfish restricted to Depths 1 through 5, orange-brown) as a function of their estimated Elo rating. Human games are grouped by Elo with vertical error bars indicating the standard error of the mean. Synthetic games are plotted with 2D error ellipses, illustrating the high, depth-dependent uncertainty inherent in mapping shallow computational searches to human Elo ratings. Dashed lines represent the linear regression fits for both populations.}}
\end{figure}

\noindent
\adamo{To disentangle the effects of raw playing strength from underlying cognitive or algorithmic differences, we compared the tension profiles of human and synthetic populations at parity of playing strength. To achieve this, we mapped the tension load of both human Rapid games and Stockfish engine matches (restricted to shallow depths from 1 to 5) onto a shared estimated Elo axis (see Fig.~\ref{fig:Skill-Calibrated}).}

\adamo{It must be explicitly stated that establishing a strict, absolute correlation between computational depth and Elo rating remains highly elusive. This mapping depends on a multitude of complex factors, including the specific engine's architecture, hardware constraints, time controls, and the nature of the evaluation function. The Elo estimates and their respective uncertainties used in our analysis were derived from the existing literature, specifically the calibration frameworks proposed by Ferreira \cite{ferreira2013impact} and the empirical experiments by Steenhuisen \cite{steenhuisen2005new}. However, because experimental conditions differ significantly across those studies, these selected values should be regarded as coarse approximations rather than exact measurements.} 

\adamo{Because the marginal gain in playing strength decays non-linearly, the uncertainty in Elo calibration is highly asymmetric and much more severe at very shallow depths. We explicitly accounted for this depth-dependent calibration uncertainty by representing the synthetic data points with 2D error ellipses.}

\adamo{Even when matched at the same estimated playing strength, the purely heuristic synthetic agent systematically exhibits a structurally higher tension load. This confirms that our metric successfully isolates intrinsic differences in the decision-making process; specifically, the AI’s tendency to maintain a more rigorous tactical posture, whereas human players, even at the same Elo level, naturally tend toward more relaxed or structurally simplified configurations.}

\section{Impact of player skill gap on strategic tension}
\label{subsec:depth_gap}
\noindent
To test the hypothesis that strategic tension is strictly dependent on the skill gap between players, we designed an experiment using simulated games with explicitly controlled asymmetries. Specifically, we employed the Stockfish engine, fixing one player's search algorithm at depth 1 to act as a proxy for a novice or quasi-random player. We then matched this baseline player against Stockfish opponents of progressively increasing strength (depths 5, 10, and 20). These asymmetric matches were generated and analyzed following the exact same procedure used for the balanced Stockfish games described in the main text. In particular, to ensure realistic starting conditions, the first ten moves of each game were initialized from common Grandmaster openings.

\begin{figure}[!htbp]
\centering
\includegraphics[width=1\linewidth]{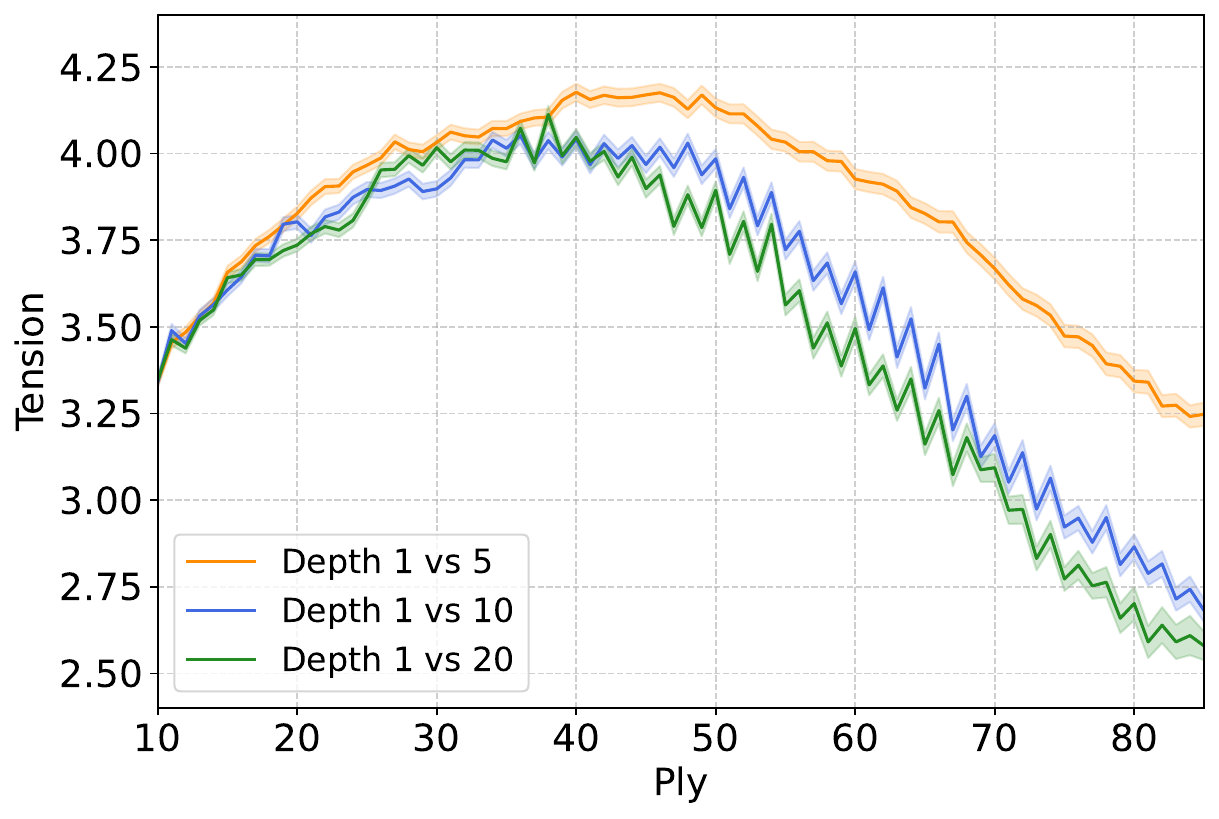}
\caption{\label{fig:depth_gap_plot} Evolution of the average tension load over time (measured in plies) for asymmetric games simulated with the Stockfish engine. The plot shows matches between a weak baseline player (capped at depth 1) and opponents at increasing computational depth: depth 5 (orange), depth 10 (blue), and depth 20 (green). Shaded areas represent the standard error. Data beyond 85 plies are omitted due to a drastic drop in available statistics; in highly asymmetric matches, the weak depth-1 player is quickly defeated and rarely survives longer games, particularly against the strongest depth-20 opponent.}
\end{figure}

The resulting tension profiles, shown in Fig.~\ref{fig:depth_gap_plot}, clearly illustrate the exact impact of the computational disparity. When the skill gap is relatively small (orange curve, depth 1 vs 5), the game still develops a moderate peak in tension. However, as the gap widens (blue curve for depth 1 vs 10, and green curve for depth 1 vs 20), the overall tension load undergoes a systematic and severe reduction. This confirms our intuition: in highly asymmetric matches, the stronger player quickly dominates and simplifies the board position, preventing the formation of the complex, sustained network of threats that typically emerges only when opponents are evenly matched.

\bibliography{references}

\end{document}